\documentclass[journal]{IEEEtran}

\PassOptionsToPackage{hyphens}{url}
\PassOptionsToPackage{bookmarks=false}{hyperref}

\IfFileExists{orcidlink.sty}{%
  \usepackage{orcidlink}
  \newcommand{\IEEEauthororcidlink}[1]{\orcidlink{##1}}%
}{%
  \newcommand{\IEEEauthororcidlink}[1]{}%
}

\usepackage[utf8]{inputenc}
\usepackage[T1]{fontenc}

\usepackage{xurl}

\usepackage{soul}
\usepackage[table]{xcolor}
\sethlcolor{yellow!45}
\soulregister\textsc7
\soulregister\textbf7
\soulregister\emph7
\soulregister\ref7
\soulregister\cite7
\soulregister\texttt7
\soulregister\footnote7
\usepackage{microtype}

\usepackage{dblfloatfix}

\usepackage{amsmath,amssymb,mathtools}
\usepackage{bm}
\usepackage{booktabs}
\usepackage{multirow}
\usepackage{array}
\usepackage{siunitx}
\usepackage{threeparttable}
\usepackage{threeparttablex}
\usepackage{tabularx}
\usepackage{makecell}
\usepackage{pbox}
\usepackage{algorithm}
\usepackage{algorithmic}

\newcolumntype{L}[1]{>{\raggedright\arraybackslash}p{#1}}
\newcolumntype{C}[1]{>{\centering\arraybackslash}p{#1}}
\newcolumntype{R}[1]{>{\raggedleft\arraybackslash}p{#1}}
\newcolumntype{Y}{>{\raggedright\arraybackslash}X}

\PassOptionsToPackage{final}{graphicx}   
\usepackage{graphicx}
\setkeys{Gin}{draft=false}

\graphicspath{{./}{./figures/}}
\DeclareGraphicsExtensions{.pdf,.png,.jpg,.jpeg}
\usepackage[caption=false,font=footnotesize]{subfig}

\usepackage[dvipsnames,table]{xcolor}

\usepackage{placeins}
\usepackage{balance}

\usepackage{cite}
\usepackage{url}

\usepackage{hyperref}

\hypersetup{
    colorlinks=true,
    linkcolor=black,
    citecolor=NavyBlue,
    urlcolor=black
}

\usepackage[capitalise,noabbrev,nameinlink]{cleveref}

\DeclareSIUnit\rpm{rpm}
\DeclareSIUnit\db{dB}
\DeclareSIUnit\fps{fps}
\DeclareSIUnit\ms{ms}

\begin{document}
\bstctlcite{IEEEexample:BSTcontrol}

\title{Fed-FSTQ: Fisher-Guided Token Quantization for Communication-Efficient Federated Fine-Tuning of LLMs on Edge Devices}

%
%
%

\author{%
  Changyu~Li$^{1}$\IEEEauthororcidlink{0009-0007-0876-0310},
  Shuanghong~Huang$^{2}$\IEEEauthororcidlink{0009-0006-0271-0040},
  Jiashen~Liu$^{3}$\IEEEauthororcidlink{0009-0000-0908-0877},
  Ming~Lei$^{1}$\IEEEauthororcidlink{0009-0007-3989-6755},
  Jidu~Xing$^{4}$\IEEEauthororcidlink{0000-0002-2679-3497},
  Kaishun~Wu$^{5}$\IEEEauthororcidlink{0000-0003-2216-0737},~\IEEEmembership{Fellow,~IEEE},
  Lu~Wang$^{6}$\IEEEauthororcidlink{0000-0001-6345-3873},~\IEEEmembership{Senior Member,~IEEE},
  and Fei~Luo$^{1}$\IEEEauthororcidlink{0000-0001-9760-1520}%
\thanks{$^{1}$Changyu~Li, Ming~Lei, and Fei~Luo are with the School of Computing and Information Technology, Great Bay University, Dongguan, China (e-mail: Changyuli.021230@gmail.com; leiming61@hrbeu.edu.cn; luofei2018@outlook.com).}%
\thanks{$^{2}$Shuanghong~Huang is with Beijing Institute of Technology, Beijing, China (e-mail: shuanghong@bit.edu.cn).}%
\thanks{$^{3}$Jiashen~Liu is with the University of Warwick, Coventry, U.K. (e-mail: jxlau1017@gmail.com).}%
\thanks{$^{4}$Jidu~Xing is with City University of Hong Kong (Dongguan), Dongguan 523808, China (e-mail: jiduo.xing@cityu-dg.edu.cn).}%
\thanks{$^{5}$Kaishun~Wu is with the Hong Kong University of Science and Technology (Guangzhou), Guangzhou, China (e-mail: wuks@hkust-gz.edu.cn).}%
\thanks{$^{6}$Lu~Wang is with the College of Computer Science and Software Engineering, Shenzhen University, Shenzhen, China (e-mail: wanglu@szu.edu.cn).}%
\thanks{This work was supported by the China NSFC Grant (U2001207, 61872248), Guangdong NSF 2017A030312008, Shenzhen Science and Technology Foundation (No. ZDSYS20190902092853047, R2020A045), the Project of DEGP (No. 2019KCXTD005), and the Guangdong ``Pearl River Talent Recruitment Program'' under Grant 2019ZT08X603. Fei~Luo is the corresponding author.}%
}

\maketitle

\begin{abstract}
Federated fine-tuning provides a practical route to adapt large language models (LLMs) on edge devices without centralizing private data. However, in mobile deployments, the training wall-clock is often dominated by straggler-limited uplink communication under heterogeneous bandwidth, intermittent participation, and non-IID client data. Although parameter-efficient fine-tuning (PEFT) methods such as LoRA and QLoRA reduce local memory and trainable parameters, repeated transmission of adapter updates remains a major bottleneck. We propose Fed-FSTQ, a {semantic-sensitivity-aware communication-control} primitive for communication-efficient federated LLM fine-tuning. Fed-FSTQ uses a lightweight token-level Fisher proxy to estimate semantic sensitivity, couples token-guided sparsification with mixed-precision adapter-update quantization, and allocates higher communication fidelity to semantically load-bearing evidence while suppressing redundant transmission. The method is drop-in compatible with standard federated PEFT pipelines and requires no change to the server aggregation rule. Experiments on multilingual QA and medical QA under non-IID partitions show that Fed-FSTQ reduces cumulative uplink traffic required to reach a fixed quality threshold by \textbf{46$\times$} relative to a Fed-LoRA baseline and improves straggler-limited wall-clock time-to-accuracy by \textbf{52\%}. Under the corrected Controlled LTE-20Mbps accounting, Fed-FSTQ reduces per-round time from 414.60s to 67.29s and reduces per-round energy from 839.20J to 146.28J, yielding a \textbf{6.16$\times$} speedup. On NVIDIA Jetson-class edge devices, Fisher-guided token reduction also yields up to a \textbf{1.55$\times$} inference speedup, demonstrating deployability under tight resource constraints.
\end{abstract}


%
\IEEEpeerreviewmaketitle

\section{Introduction}
\label{sec:introduction}

\IEEEPARstart{T}{he} deployment of Large Language Models (LLMs) on mobile and edge devices is reshaping personalized computing, enabling privacy-preserving clinical assistants, multilingual copilots, and real-time on-device code generation~\cite{brown2020gpt3,touvron2023llama,devlin2019bert,vaswani2017attention,shi2016edge}. Federated Learning (FL) provides a natural mechanism for adapting such models without centralizing sensitive user data~\cite{mcm2017fedavg,kairouz2021flsurvey,li2020flspm}. In practical deployments, secure aggregation or differential privacy may further tighten the effective communication budget~\cite{bonawitz2017secureagg,abadi2016dpsgd}. However, scaling LLM adaptation to cross-device settings faces a persistent systems bottleneck: edge accelerators continue to improve local computation, while wireless uplinks remain scarce, time-varying, and highly heterogeneous~\cite{lim2020fedcomst,bonawitz2019flsystems}.

\smallskip
\noindent
\emph{The uplink wall in federated LLM fine-tuning.}
In synchronous federated fine-tuning, each round is often governed by the slowest participating clients, making the training wall-clock straggler-limited~\cite{bonawitz2019flsystems}. Parameter-Efficient Fine-Tuning (PEFT), including LoRA and QLoRA, substantially reduces trainable parameters and local memory, but it does not eliminate repeated uplink transmission of adapter updates across communication rounds~\cite{hu2022lora,dettmers2023qlora}. Under Non-IID client data, adapter updates also become statistically heterogeneous, which can amplify client drift and make communication-efficient training more brittle~\cite{li2020fedprox,wang2020fednova,karimireddy2020scaffold,reddi2021fedopt,acar2021feddyn}. Thus, for mobile and edge FL, reducing the number of trainable parameters is not sufficient; the transmitted update itself must become more information-efficient.

\smallskip
\noindent
\emph{Why parameter-centric compression is insufficient.}
A standard remedy is to compress updates using quantization, sparsification, periodic averaging, or error-feedback~\cite{alistarh2017qsgd,lin2018dgc,reisizadeh2020fedpaq,karimireddy2019errorfeedback}. These methods are effective at reducing payload size, but they are largely parameter-centric: coordinates are selected or quantized using uniform, stochastic, magnitude-based, or low-rank rules. Such policies do not explicitly model token-level semantic sensitivity. This matters for LLM adaptation because correctness can depend on rare but decisive tokens, such as clinical negations, domain entities, code delimiters, or multilingual morphology. Attention-driven token reduction methods such as DynamicViT or ToMe provide useful adaptive-computation heuristics~\cite{rao2021dynamicvit,bolya2022tome}, but directly applying them to text can be unreliable because low-attention tokens are not necessarily semantically redundant.

\smallskip
\noindent
\emph{Our approach: Fisher-guided communication control.}
We propose Fed-FSTQ (\underline{F}isher-\underline{S}pectrum-aware \underline{T}oken \underline{Q}uantization), a communication-control primitive for federated LLM fine-tuning. Instead of treating all transmitted coordinates as equally valuable, Fed-FSTQ uses a lightweight token-level Fisher proxy to estimate which token evidence is most consequential for the local loss geometry. It then couples token-guided sparsification with mixed-precision adapter-update quantization, assigning higher precision to token-conditioned update directions that carry larger Fisher-weighted sensitivity and pruning or coarsely quantizing low-sensitivity components. This converts semantic sensitivity into a systems control signal: scarce uplink bits are spent on the evidence most likely to preserve downstream quality under Non-IID heterogeneity.

\smallskip
\noindent
\emph{Drop-in deployability and privacy-aware scope.}
Fed-FSTQ is designed as a drop-in module for standard federated PEFT pipelines such as FedAvg+LoRA~\cite{mcm2017fedavg,hu2022lora}. The server aggregation rule remains unchanged: clients transmit compressed sparse adapter updates, and the server dequantizes and aggregates them using FedAvg-compatible weighted averaging. For privacy-sensitive deployments, Fed-FSTQ should be integrated with secure aggregation or differential privacy under fixed-shape or metadata-protected packing; raw sparse supports, bit-width tags, and Fisher-derived masks should not be exposed directly. Thus, Fed-FSTQ is a communication-control mechanism rather than a standalone privacy guarantee.

\smallskip
\noindent
\emph{Results and contributions.}
We evaluate Fed-FSTQ on multilingual QA (Fed-Aya) and medical QA (Fed-Med)~\cite{jin2019pubmedqa,jin2020medqa} using a virtual edge testbed that combines measured Jetson computation with emulated heterogeneous LTE uplinks. Fed-FSTQ reduces cumulative uplink traffic required to reach a fixed quality threshold by \textbf{46$\times$} relative to Fed-LoRA and improves straggler-limited wall-clock time-to-accuracy by \textbf{52\%}. Under the corrected Controlled LTE-20Mbps accounting, it reduces per-round time from 414.60s to 67.29s and reduces per-round energy from 839.20J to 146.28J, corresponding to a \textbf{6.16$\times$} speedup. On NVIDIA Jetson-class devices, the learned Fisher-guided masks also enable up to a \textbf{1.55$\times$} inference speedup.

The main contributions are:
\begin{itemize}
    \item \emph{Fisher-guided federated communication control.} We formulate token-level Fisher sensitivity as a communication-control primitive for federated LLM fine-tuning, enabling mixed-precision transmission that preserves semantically decisive evidence under severe uplink constraints.
    \item \emph{Token-to-parameter coupling for training-time compression.} We show how token-level sensitivity enters the token-filtered local objective and induces parameter-level Fisher statistics for adapter-coordinate sparsification and bit allocation.
    \item \emph{System-level gains under heterogeneity.} We demonstrate substantial reductions in cumulative uplink traffic, per-round latency, energy, and straggler-limited time-to-accuracy under Non-IID client distributions and heterogeneous mobile uplinks.
    \item \emph{Edge deployability and complementary benefits.} We validate feasibility on NVIDIA Jetson-class hardware and show that Fisher-guided token selection benefits both training communication and inference efficiency.
\end{itemize}

\section{Related Work}
\label{sec:related}

\subsection{Communication-Efficient Federated Learning}
Communication is a first-order systems bottleneck in federated learning (FL)~\cite{mcm2017fedavg,kairouz2021flsurvey,li2020flspm}, particularly in mobile and edge deployments where uplinks are heterogeneous, participation is intermittent, and synchronous rounds become straggler-limited~\cite{lim2020fedcomst,bonawitz2019flsystems,shi2016edge}. This view is reinforced by early system and optimization analyses emphasizing constraints beyond the datacenter~\cite{konecny2016fedoptbeyond,konecny2016comm}, and is operationalized in benchmarks and stacks such as LEAF, FedScale, Flower, and TensorFlow Federated~\cite{caldas2018leaf,hsieh2021fedscale,beutel2020flower}. When privacy requirements are stringent, pairing FL with secure aggregation or differential privacy further tightens the effective communication budget~\cite{bonawitz2017secureagg,abadi2016dpsgd}.

To mitigate uplink costs, a large body of gradient compression methods has been proposed. \textbf{Quantization} approaches (e.g., QSGD, TernGrad, 1-bit SGD, signSGD) reduce update precision, often with convergence guarantees under standard stochastic assumptions~\cite{alistarh2017qsgd,wen2017terngrad,seide20141bit,bernstein2018signsgd}. Recent work also revisits \emph{binarization} as a learnable compression primitive. FedBAT performs binarization-aware local training to reduce approximation error relative to post-hoc binarization, improving communication efficiency while maintaining accuracy~\cite{li2024fedbat}. \textbf{Error-feedback} corrects compression bias and stabilizes convergence in practice~\cite{karimireddy2019errorfeedback}. \textbf{Sparsification} communicates only a subset of significant coordinates (e.g., top-$k$), exploiting empirical gradient sparsity~\cite{lin2018dgc,stich2019sparsgd,sattler2019stc}. \textbf{Structured compression} further leverages low-rank approximations to reduce payloads~\cite{vogels2019powersgd}. In FL, FedPAQ combines periodic averaging with quantization to reduce both communication frequency and payload size~\cite{reisizadeh2020fedpaq}.

However, scaling parameter-centric schemes to federated LLM fine-tuning exposes practical limitations. First, even aggressively compressed updates can accumulate to prohibitive traffic over many rounds under straggler-limited synchronization~\cite{bonawitz2019flsystems}. Second, many schemes remain largely uniform or magnitude-driven and do not explicitly model \emph{semantic sensitivity} at the token level; in non-IID regimes, this can disproportionately attenuate rare but task-critical signals~\cite{li2020fedprox}. Fed-FSTQ complements this literature by introducing a data-centric control primitive that prioritizes semantically consequential evidence before allocating communication fidelity, thereby improving straggler-limited time-to-accuracy.
\subsection{Federated Fine-Tuning of Large Language Models}

Parameter-efficient fine-tuning (PEFT) is the dominant paradigm for adapting LLMs on resource-constrained devices.
Techniques such as adapters~\cite{houlsby2019adapters}, prefix-tuning~\cite{li2021prefix}, prompt tuning~\cite{lester2021prompttuning}, and LoRA~\cite{hu2022lora} freeze the backbone and train only a small set of parameters, substantially reducing local compute and memory.
Adaptive variants such as AdaLoRA allocate low-rank capacity under a constrained budget~\cite{zhang2023adaptive}.
Quantized fine-tuning methods such as QLoRA reduce memory pressure further by quantizing the frozen backbone while training low-rank adapters~\cite{dettmers2023qlora}.

While PEFT makes on-device training feasible, it does not remove the \emph{uplink bottleneck}.
Systems evidence indicates that communication often dominates end-to-end latency in mobile deployments due to bandwidth variability and intermittent connectivity~\cite{bonawitz2019flsystems,shi2016edge,lim2020fedcomst}.
Moreover, non-IID distributions exacerbate client drift and instability, motivating robust optimization corrections such as FedProx, SCAFFOLD, FedNova, FedDyn, and adaptive federated optimization~\cite{li2020fedprox,karimireddy2020scaffold,wang2020fednova,acar2021feddyn,reddi2021fedopt}.
Unlike methods that primarily reduce the \emph{number of trainable parameters}, Fed-FSTQ targets the \emph{information density} of the transmitted update by prioritizing token-level evidence and allocating mixed precision so that scarce uplink capacity is spent on updates most consequential for downstream correctness.

Recent FedPEFT systems address complementary aspects of federated LLM adaptation.
AdaFL/FedAdapter reduces client-side training cost through adapter-based efficient federated NLP~\cite{cai2023fedadapter};
LEGEND adapts LoRA depth and rank distribution for heterogeneous devices~\cite{liu2025legend};
and FedLoDrop introduces dropout-style LoRA regularization for generalized federated LLM fine-tuning~\cite{xie2026fedlodrop}.
These methods mainly decide which computation, layer, rank, or adapter structure should be used, whereas Fed-FSTQ decides which token-conditioned adapter-update information should be transmitted and at what precision. Accordingly, Fed-FSTQ is complementary to scheduling-, structure-, and regularization-based FedPEFT methods and can be composed with them.

Relatedly, LLM quantization has been extensively studied for efficient deployment, including integer-matrix kernels and post-training quantization methods such as LLM.int8(), GPTQ, SmoothQuant, AWQ, and ZeroQuant~\cite{dettmers2022llmint8,frantar2022gptq,xiao2023smoothquant,lin2023awq,yao2022zeroquant}.
These works primarily target inference-time acceleration and memory reduction, whereas Fed-FSTQ focuses on federated uplink efficiency under repeated update exchange.

\subsection{Curvature-Aware Optimization and Compression}
Fisher information and other curvature-aware signals provide a principled alternative to magnitude heuristics for modeling saliency. In optimization, natural gradient methods and scalable curvature approximations such as K-FAC exploit local curvature structure to improve convergence~\cite{amari1998natural,martens2015kfac}. In model compression, classic second-order analyses (e.g., Optimal Brain Damage and Optimal Brain Surgeon) use Hessian-based approximations to identify prunable parameters with minimal loss increase~\cite{lecun1989obd,hassibi1993obs}. Fisher information has also served as a sensitivity proxy in continual learning (e.g., EWC) to protect critical weights under nonstationarity~\cite{kirkpatrick2017ewc}. More recent pruning work emphasizes preserving gradient flow or sensitivity at initialization (e.g., SNIP and GraSP)~\cite{lee2019snip,wang2020grasp}. Complementary lines study end-to-end compression pipelines that combine pruning and quantization~\cite{han2016deepcompression,jacob2018quant}, sparsity induction via stochastic regularization~\cite{molchanov2017variationaldropout}, and structural perspectives such as the lottery ticket hypothesis~\cite{frankle2019lottery}.

Fed-FSTQ draws inspiration from this tradition but pivots the application domain in two key ways: it targets \textbf{federated uplink efficiency} rather than inference-time compression, and it lifts sensitivity modeling from the parameter level to the \textbf{token level}. In this context, Fed-FSTQ uses Fisher-guided sensitivity not merely as a static pruning criterion, but as a dynamic rate--distortion control signal that modulates communication fidelity under client and network heterogeneity.

\subsection{Token Pruning and Adaptive Computation}
Adaptive computation has been widely explored in Transformer architectures~\cite{vaswani2017attention}, especially in Vision Transformers where patch tokens are naturally discrete~\cite{dosovitskiy2021vit}. Methods such as DynamicViT, TokenLearner, ATS, A-ViT, EViT, and token merging (ToMe) improve inference speed by dropping or merging redundant patches using attention- or confidence-driven heuristics~\cite{rao2021dynamicvit,ryoo2021tokenlearner,fayyaz2022ats,yin2022avit,liang2022evit,bolya2022tome}.

However, directly transplanting these heuristics to federated LLM fine-tuning is non-trivial. Unlike image patches, text tokens exhibit high \textbf{structural density}; tokens that appear low-attention (e.g., negations in clinical text or delimiters in code) can be pivotal for correctness. Under non-IID heterogeneity, inconsistent token dropping or coarse treatment of such tokens can degrade semantic reliability and compound client drift~\cite{li2020fedprox}. Fed-FSTQ departs from activation-driven heuristics by adopting a \textbf{Fisher-guided} sensitivity signal. Using Fisher information as a proxy for how token evidence shapes the loss landscape, it prioritizes and allocates precision to tokens that are most consequential, preserving structural integrity under high compression ratios.

\smallskip
\noindent\textbf{Broader Context on Reliable Edge Systems.}
Our focus on preserving rare but safety-critical evidence under constraints resonates with broader trends in reliability-aware edge analytics under nonstationarity~\cite{li2026pgtmt}. Fed-FSTQ extends this philosophy to federated LLM adaptation by treating communication control as a sensitivity-aware primitive for robust learning in heterogeneous networks.


\section{Problem Formulation and System Model}
\label{sec:problem}

\subsection{Federated Low-Rank Adaptation (Fed-LoRA)}
We consider a cross-device federated system with a central server and $K$ heterogeneous edge clients (e.g., mobile SoCs and Jetson-class modules), indexed by $\mathcal{C}=\{1,\dots,K\}$~\cite{mcm2017fedavg,kairouz2021flsurvey,bonawitz2019flsystems,shi2016edge}. Client $k$ holds a private dataset $\mathcal{D}_k=\{(x_i,y_i)\}_{i=1}^{N_k}$ drawn from a client-specific distribution $\mathcal{P}_k$, capturing the statistical heterogeneity (non-IID) endemic to practical deployments~\cite{kairouz2021flsurvey,li2020flspm,li2020fedprox}. At communication round $t$, the server samples a subset $\mathcal{S}_t \subseteq \mathcal{C}$ (partial participation), and each selected client performs local PEFT updates; such sampling and heterogeneity are widely studied and benchmarked in real FL stacks (e.g., LEAF, FedScale, Flower, TFF)~\cite{caldas2018leaf,hsieh2021fedscale,beutel2020flower}.

We adopt Low-Rank Adaptation (LoRA)~\cite{hu2022lora}. For Transformer layer $l\in\{1,\dots,L\}$ with frozen weights $\mathbf{W}_0^{(l)}\in\mathbb{R}^{d_{\text{out}}\times d_{\text{in}}}$, LoRA injects trainable low-rank matrices $\mathbf{B}^{(l)}\in\mathbb{R}^{d_{\text{out}}\times r}$ and $\mathbf{A}^{(l)}\in\mathbb{R}^{r\times d_{\text{in}}}$ with $r\ll \min(d_{\text{out}},d_{\text{in}})$. For input activation $\mathbf{h}_{\text{in}}^{(l)}$, the layer output is
\begin{equation}
\mathbf{h}_{\text{out}}^{(l)}=\mathbf{W}_0^{(l)}\mathbf{h}_{\text{in}}^{(l)}+\frac{\alpha_{\text{lora}}}{r}\mathbf{B}^{(l)}\mathbf{A}^{(l)}\mathbf{h}_{\text{in}}^{(l)},
\label{eq:lora_forward}
\end{equation}
where $\alpha_{\text{lora}}$ is the LoRA scaling. Let $\boldsymbol{\Theta}\triangleq\{\mathbf{A}^{(l)},\mathbf{B}^{(l)}\}_{l=1}^L$ denote all trainable adapter parameters (while the backbone is frozen, optionally quantized as in QLoRA~\cite{dettmers2023qlora}). The global objective is the weighted empirical risk
\begin{equation}
\min_{\boldsymbol{\Theta}}~\mathcal{F}(\boldsymbol{\Theta}) \triangleq \sum_{k=1}^{K}\frac{N_k}{N}\,\mathcal{L}_k(\boldsymbol{\Theta};\mathcal{D}_k),
\label{eq:fed_objective}
\end{equation}
where $N=\sum_k N_k$ and $\mathcal{L}_k$ is the local causal LM loss (e.g., next-token cross-entropy) instantiated on Transformer language modeling objectives~\cite{vaswani2017attention,brown2020gpt3,touvron2023llama,devlin2019bert}.

\paragraph*{Local update and aggregation (FedAvg-compatible)}
At round $t$, each participating client initializes $\boldsymbol{\Theta}_{k,t}^{(0)}\gets \boldsymbol{\Theta}_t$ and runs $E$ local steps:
\begin{equation}
\boldsymbol{\Theta}_{k,t}^{(\tau+1)}=\boldsymbol{\Theta}_{k,t}^{(\tau)}-\eta\,\widehat{\nabla}\mathcal{L}_k\!\left(\boldsymbol{\Theta}_{k,t}^{(\tau)};\xi_{k,t}^{(\tau)}\right),
\qquad \tau=0,\dots,E-1,
\label{eq:local_sgd}
\end{equation}
where $\xi_{k,t}^{(\tau)}$ denotes a local minibatch and $\widehat{\nabla}$ is a stochastic gradient estimator. The (uncompressed) model delta is
\begin{equation}
\Delta\boldsymbol{\Theta}_{k,t}\triangleq \boldsymbol{\Theta}_{k,t}^{(E)}-\boldsymbol{\Theta}_t.
\label{eq:model_delta}
\end{equation}
The server updates via a FedAvg-style weighted aggregation~\cite{mcm2017fedavg}:
\begin{equation}
\begin{split}
\boldsymbol{\Theta}_{t+1} &= \boldsymbol{\Theta}_t + \sum_{k\in\mathcal{S}_t:a_{k,t}=1} w_{k,t}\,\widetilde{\Delta\boldsymbol{\Theta}}_{k,t}, \\
w_{k,t} &\triangleq \frac{N_k}{\sum_{j\in\mathcal{S}_t:a_{j,t}=1}N_j}.
\end{split}
\label{eq:server_agg}
\end{equation}


where $\widetilde{\Delta\boldsymbol{\Theta}}_{k,t}$ is the decompressed client update and $a_{k,t}$ is an availability indicator (defined below). This formulation remains compatible with FedOpt-style server optimizers and heterogeneity-aware corrections (e.g., FedProx, SCAFFOLD, FedNova, FedDyn)~\cite{reddi2021fedopt,li2020fedprox,karimireddy2020scaffold,wang2020fednova,acar2021feddyn}.
\subsection{Stochastic Uplink Channel and Straggler-Limited Latency}
Mobile and edge FL is often constrained by uplink communication under bandwidth heterogeneity and intermittent participation~\cite{lim2020fedcomst,bonawitz2019flsystems,shi2016edge}. 
We study synchronous rounds because cross-device FL deployments are often straggler-limited: the round completion time is governed by the slowest available client rather than the average client~\cite{bonawitz2019flsystems,hsieh2021fedscale}.

At round $t$, each participating client $k\in\mathcal{S}_t$ transmits a compressed message
\begin{equation}
\mathbf{m}_{k,t} \triangleq \mathrm{Enc}\!\left(\mathcal{Q}_{k,t}(\Delta\boldsymbol{\Theta}_{k,t})\right),
\end{equation}
where $\mathcal{Q}_{k,t}$ denotes (possibly sparse, mixed-precision) quantization/compression, and $\mathrm{Enc}(\cdot)$ packs values together with required metadata (e.g., masks/indices, quantization scales, bit-width tags)~\cite{konecny2016comm,alistarh2017qsgd,lin2018dgc,reisizadeh2020fedpaq}. Let $\mathrm{bits}(\mathbf{m}_{k,t})$ denote the payload length. A convenient decomposition that separates structure and value costs is
\begin{equation}
{
\mathrm{bits}(\mathbf{m}_{k,t})
~=~
\underbrace{\mathrm{bits}(\mathcal{I}_{k,t})}_{\text{indices/mask}}
~+~
\underbrace{\sum_{j\in\mathcal{I}_{k,t}} b_{k,t}(j)}_{\text{quantized values}}
~+~
\underbrace{\mathrm{bits}(\text{side info})}_{\text{scales/tags}},
}
\label{eq:payload_decomp}
\end{equation}
where $\mathcal{I}_{k,t}$ is the set of transmitted coordinates and $b_{k,t}(j)\in\mathbb{Z}_{\ge 0}$ is the bit-width assigned to coordinate $j$ (with $b_{k,t}(j)=0$ implying pruning). Eq.~\eqref{eq:payload_decomp} covers uniform quantization (fixed $b$), top-$k$ sparsification, and mixed-precision signaling, and it is compatible with bias-correction mechanisms such as error-feedback and memory~\cite{karimireddy2019errorfeedback,stich2019sparsgd}.

Importantly, Eq.~\eqref{eq:payload_decomp} is used for all reported payloads and tables; we never report value-only payloads. This is necessary because, under aggressive sparsity, index or mask metadata can dominate the quantized-value payload. Therefore, Fed-FSTQ accepts a support and bit-width allocation only when the full packed message, including indices/masks, bit-width tags, group-wise scales, and side information, satisfies the target communication budget.

\paragraph*{Throughput and intermittent participation}
Let $R_{k,t}$ be the effective uplink throughput available to client $k$ at round $t$. We treat $\{R_{k,t}\}_t$ as a stochastic process capturing time-varying cellular/WiFi conditions and cross-traffic:
\begin{equation}
{
T_{\mathrm{comm}}^{(k,t)}
~=~
\frac{\mathrm{bits}(\mathbf{m}_{k,t})}{R_{k,t}},
\qquad
R_{k,t}\sim \mathcal{P}^{(k)}_{\mathrm{bw}}.
}
\label{eq:comm_time}
\end{equation}
We model intermittent participation with an availability indicator $a_{k,t}\in\{0,1\}$ (temporary offline or dropped round), consistent with cross-device deployments~\cite{bonawitz2019flsystems,lim2020fedcomst}. This stochastic-uplink model separates the algorithmic payload decision $\mathrm{bits}(\mathbf{m}_{k,t})$ from the time-varying wireless condition $R_{k,t}$, allowing us to evaluate both controlled LTE-20Mbps accounting and heterogeneous straggler-limited uplink regimes.

\paragraph*{Network profiles used in evaluation (disambiguating ``4G LTE'')}
To avoid conflating distinct uplink settings under a single label, we use two explicit LTE profiles throughout the paper and refer to them by name in all table and figure captions. Profile A fixes a controlled rate to isolate payload effects in per-round breakdown analysis. Profile B draws $R_{k,t}$ from a client-specific distribution with a slow tail to model stragglers in end-to-end time-to-accuracy experiments. Under Profile A, $T_{\mathrm{comm}}^{(k,t)}$ follows directly from Eq.~\eqref{eq:comm_time} with a fixed rate, while under Profile B the synchronous latency becomes straggler-dominated through Eq.~\eqref{eq:straggler_model}.

\paragraph*{Straggler-limited round time}
In synchronous FL, the per-round wall-clock latency is
\begin{equation}
T_{\mathrm{round}}^{(t)}
~=~
T_{\mathrm{srv}}^{(t)}+
\max_{k\in\mathcal{S}_t:~a_{k,t}=1}
\left(
T_{\mathrm{comp}}^{(k,t)} + \frac{\mathrm{bits}(\mathbf{m}_{k,t})}{R_{k,t}}
\right),
\label{eq:straggler_model}
\end{equation}
where $T_{\mathrm{comp}}^{(k,t)}$ is local computation time and $T_{\mathrm{srv}}^{(t)}$ accounts for server-side aggregation overhead~\cite{bonawitz2019flsystems,hsieh2021fedscale}. Eq.~\eqref{eq:straggler_model} makes the key systems lever explicit: under heterogeneous uplinks with a slow tail, reducing $\mathrm{bits}(\mathbf{m}_{k,t})$ improves both average and tail round latency, and therefore time-to-accuracy. This remains true when local fine-tuning is made feasible by PEFT/QLoRA~\cite{hu2022lora,dettmers2023qlora}, and when privacy mechanisms further tighten effective communication budgets~\cite{bonawitz2017secureagg,abadi2016dpsgd}.

\subsection{Problem Statement: Fisher-Weighted Rate--Distortion Optimization}
\label{sec:problem_rdo}

Communication-efficient FL methods usually compress adapter updates in a parameter-centric way, using uniform quantization, magnitude sparsification, periodic averaging, or error feedback~\cite{konecny2016comm,alistarh2017qsgd,lin2018dgc,reisizadeh2020fedpaq,karimireddy2019errorfeedback}. This is brittle for language workloads: correctness may depend on rare but decisive tokens such as clinical negations, domain entities, or code delimiters, while simple coordinate magnitude or attention scores do not reliably measure their importance. Under Non-IID client data, indiscriminate compression can also amplify client drift by suppressing long-tail evidence~\cite{li2020fedprox,kairouz2021flsurvey}.

We therefore use a tractable \emph{Fisher-weighted rate--distortion surrogate} for communication control. The goal is not to claim a full information-geometric characterization of the LLM loss landscape, but to obtain a separable sensitivity-aware objective that can be implemented on edge clients with diagonal empirical Fisher statistics. The token-level Fisher proxy shapes these adapter-coordinate statistics through the token-filtered local objective derived in Sec.~\ref{sec:coupling}.

\paragraph*{Fisher-weighted distortion}
Let $\hat{\mathbf{F}}_t$ be a diagonal empirical Fisher approximation at round $t$. For an update $\Delta\boldsymbol{\Theta}_{k,t}$ and its compressed counterpart $\widetilde{\Delta\boldsymbol{\Theta}}_{k,t}$, define
\begin{equation}
\begin{split}
D_{\hat{\mathbf{F}}_t}\!\left(\Delta\boldsymbol{\Theta}_{k,t},\widetilde{\Delta\boldsymbol{\Theta}}_{k,t}\right)
&\triangleq \mathbf{e}_{k,t}^{\top}\hat{\mathbf{F}}_t\mathbf{e}_{k,t}, \\
\mathbf{e}_{k,t}&\triangleq \Delta\boldsymbol{\Theta}_{k,t}-\widetilde{\Delta\boldsymbol{\Theta}}_{k,t}.
\end{split}
\label{eq:fisher_distortion}
\end{equation}
Under the diagonal surrogate,
\begin{equation}
D_{\hat{\mathbf{F}}_t}=\sum_j \hat{F}_t(j)\,\big(\Delta\theta_{k,t}(j)-\widetilde{\Delta\theta}_{k,t}(j)\big)^2.
\label{eq:fisher_distortion_diag}
\end{equation}
Thus, compression error is penalized more strongly on high-sensitivity coordinates, consistent with classical second-order saliency and modern sensitivity-based pruning principles~\cite{lecun1989obd,hassibi1993obs,lee2019snip,wang2020grasp}.

\paragraph*{Rate--distortion objective}
We minimize expected uplink payload subject to bounded Fisher-weighted distortion:
\begin{align}
\min_{\pi}~~& \mathbb{E}\big[\mathrm{bits}(\mathbf{m}_{k,t})\big] \nonumber\\
\mathrm{s.t.}~~& \mathbb{E}\!\left[D_{\hat{\mathbf{F}}_t}\!\left(\Delta\boldsymbol{\Theta}_{k,t},\widetilde{\Delta\boldsymbol{\Theta}}_{k,t}\right)\right]\le \epsilon,
\label{eq:rdo_constraint}
\end{align}
where $\pi$ maps a local update and system state to a sparse support and mixed-precision allocation. Equivalently, using a Lagrangian relaxation,
\begin{equation}
\min_{\pi}~\mathbb{E}\!\left[\mathrm{bits}(\mathbf{m}_{k,t})+\lambda D_{\hat{\mathbf{F}}_t}\!\left(\Delta\boldsymbol{\Theta}_{k,t},\widetilde{\Delta\boldsymbol{\Theta}}_{k,t}\right)\right].
\label{eq:rdo_lagrangian}
\end{equation}

\paragraph*{Separable allocation}
Combining the diagonal distortion in Eq.~\eqref{eq:fisher_distortion_diag} with the sparse payload decomposition in Eq.~\eqref{eq:payload_decomp} gives
\begin{equation}
\label{eq:rdo_separable}
\begin{split}
\min_{\{\mathcal{I}_{k,t},b_{k,t}(j)\}}~~&
\mathrm{bits}(\mathcal{I}_{k,t})+\mathrm{bits}(\mathrm{side~info}) \\
&+\sum_{j\in\mathcal{I}_{k,t}}\!\left[b_{k,t}(j)+\lambda\hat{F}_t(j)\mathbb{E}\big[e_{k,t}(j;b_{k,t}(j))^2\big]\right],
\end{split}
\end{equation}
where $b_{k,t}(j)=0$ denotes pruning and $e_{k,t}(j;b)$ is the coordinate-wise quantization error. This form turns the communication decision into scalar bit-allocation choices plus support metadata. Increasing precision from $b$ to $b+\Delta b$ is justified only when
\begin{equation}
\lambda\hat{F}_t(j)
\frac{\mathbb{E}[e_{k,t}(j;b)^2]-\mathbb{E}[e_{k,t}(j;b+\Delta b)^2]}{\Delta b}\ge 1.
\label{eq:marginal_condition}
\end{equation}
Hence, Fisher-critical coordinates receive higher precision, while low-sensitivity coordinates are coarsely quantized or pruned. Through Eq.~\eqref{eq:payload_decomp}--\eqref{eq:straggler_model}, this directly targets straggler-limited latency without uniformly distorting optimization-critical directions.

\paragraph*{Approximation with token-level sensitivity}
Fed-FSTQ approximates this surrogate without forming a full Fisher matrix. It uses squared gradients with respect to token embeddings as a lightweight token-level Fisher proxy, then lets the token-filtered local objective induce the parameter-level Fisher statistics used for support selection and mixed precision. Thus, token sensitivity does not replace the parameter-level objective; it shapes it through training gradients, as made explicit in Sec.~\ref{sec:coupling}.

\begin{figure*}[!t]
\centering
\includegraphics[width=6.8in]{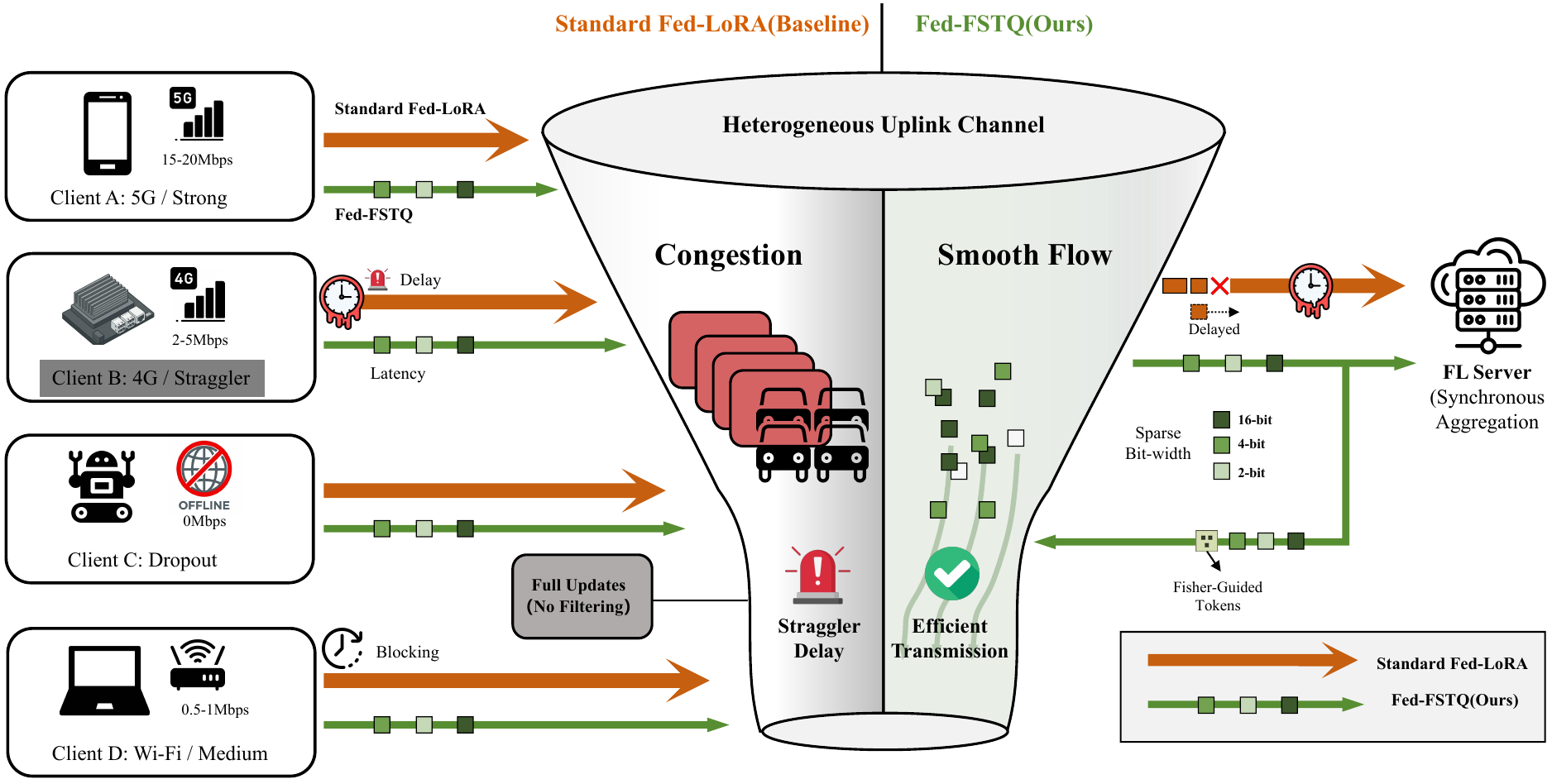}
\caption{\textbf{The Uplink Bottleneck in Federated LLM Fine-Tuning.} 
Under stochastic channel conditions ($R_{k,t}$), standard Fed-LoRA (red arrows, dense blocks) suffers from straggler delays per Eq.~\eqref{eq:straggler_model}.
Fed-FSTQ (green arrows, sparse blocks) reduces $\mathrm{bits}(\mathbf{m}_{k,t})$ via Fisher-guided semantic compression, enabling efficient transmission even under constrained and heterogeneous uplinks. 
The straggler client, highlighted with a clock icon, dominates the round completion time in the baseline scenario, while Fed-FSTQ mitigates this bottleneck through aggressive yet semantically-aware compression.}
\label{fig:uplink_bottleneck}
\end{figure*}


\section{Fed-FSTQ: Methodology and System Design}
\label{sec:method}

\begin{figure}[!t]
\centering
\includegraphics[width=3.4in]{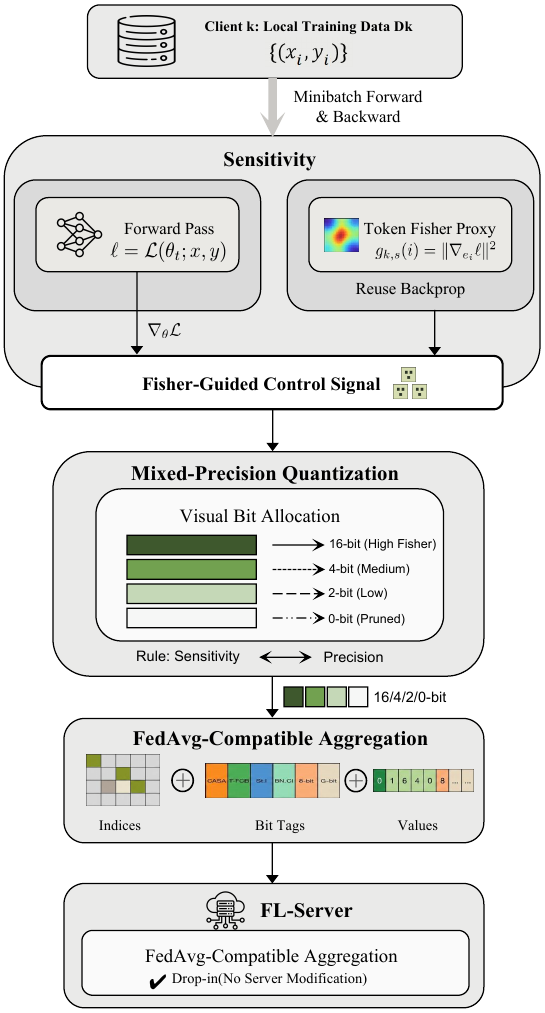}
\caption{\textbf{System Architecture of Fed-FSTQ.} {Fed-FSTQ decouples transmission fidelity from parameter magnitude by allocating bits according to Fisher-guided sensitivity.
\textbf{(1) Sensitivity estimation:} During standard backpropagation, each client computes squared gradients w.r.t. input embeddings as a token-level Fisher proxy~\cite{martens2015kfac}.
\textbf{(2) Mixed-precision allocation:} A Fisher-weighted rate--distortion policy assigns discrete bit-widths, e.g., 0/2/4/16-bit, to update coordinates, preserving load-bearing evidence while pruning low-impact components.
\textbf{(3) Sparse uplink \& aggregation:} Clients transmit a packed sparse message, including indices or masks, bit-tags, and values; the server dequantizes and aggregates via FedAvg-compatible weighted averaging.}}
\label{fig:system_overview}
\end{figure}

We now present Fed-FSTQ, a system primitive that operationalizes the Fisher-weighted rate--distortion surrogate in Sec.~\ref{sec:problem_rdo}.
As shown in Fig.~\ref{fig:system_overview}, Fed-FSTQ sits between the local PEFT training loop and the bandwidth-limited uplink, acting as a \emph{semantic reliability gate} that (i) estimates which token-level evidence is load-bearing for the local loss geometry, and (ii) allocates transmission bits accordingly.
In contrast to activation- or attention-only pruning heuristics~\cite{liang2022evit,bolya2022tome}, Fed-FSTQ uses a lightweight Fisher proxy to estimate token sensitivity, and couples this signal with mixed-precision quantization and sparse message packing that remains compatible with standard FedAvg aggregation~\cite{mcm2017fedavg}.
The pipeline consists of three stages: \textbf{(1) Fisher-proxy sensitivity estimation}, \textbf{(2) Fisher-weighted rate--distortion bit allocation}, and \textbf{(3) sparse uplink and aggregation}.

\subsection{Stage 1: Token-Level Fisher-Proxy Sensitivity}
\label{sec:fisher_sensitivity}

In non-IID federated fine-tuning, a client update intermixes transferable signal and locally idiosyncratic variation.
Fed-FSTQ extracts a sensitivity signal directly from the local loss geometry using a Fisher proxy, but does so in a token-centric and edge-feasible manner.

\paragraph{Token-level sensitivity}
Consider a training sequence of length $L$ with tokens $\{t_i\}_{i=1}^{L}$ and embeddings $\mathbf{e}_i\in\mathbb{R}^{d_e}$.
At client $k$ and local step $s$, we define the instantaneous token sensitivity as
\begin{equation}
g_{k,s}(i)\triangleq \left\|\nabla_{\mathbf{e}_i}\mathcal{L}_{k,s}\right\|_2^2,
\label{eq:token_grad_sq}
\end{equation}
where $\mathcal{L}_{k,s}$ is the local loss for the minibatch at step $s$.
This quantity is an empirical Fisher proxy in embedding space: it measures how strongly the loss reacts to perturbations of token $t_i$ at the current iterate, and therefore highlights structurally decisive evidence, such as clinical negations or code delimiters, that attention magnitude can underweight.

\paragraph{Temporal smoothing for variance control}
Because $g_{k,s}(i)$ is noisy under minibatch SGD, we apply an exponential moving average (EMA),
\begin{equation}
S_{k,s}(i)=\rho\,S_{k,s-1}(i)+(1-\rho)\,g_{k,s}(i),
\qquad \rho\in(0,1),
\label{eq:token_ema}
\end{equation}
with $\rho=0.9$ by default.
The EMA stabilizes token ranking under stochastic gradients and provides a stable control signal for subsequent support selection and bit allocation.
Empirically, this stabilizing effect aligns with the observed increase in Token Recall in Sec.~\ref{sec:eval_reliability}, consistent with Fisher-guided pruning acting as a semantic denoiser rather than a purely lossy compressor.

\begin{figure*}[!t]
\centering
\includegraphics[width=6.8in]{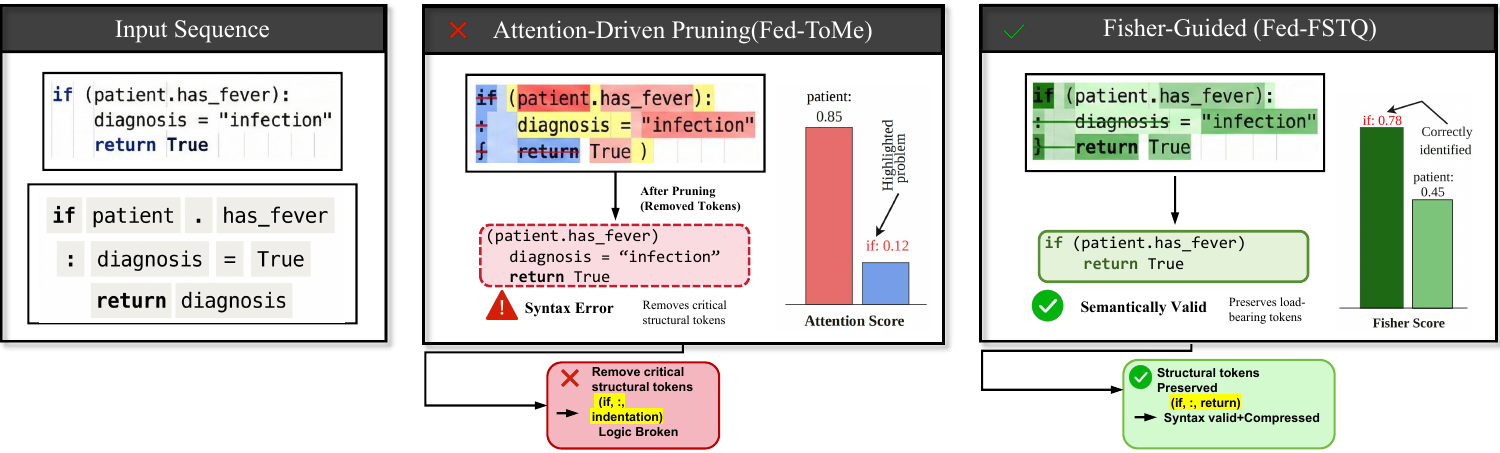}
\caption{\textbf{Fisher vs.\ Attention Heatmap.} {Attention may emphasize high-frequency connectors, whereas the Fisher proxy highlights structurally decisive tokens whose removal breaks logical validity, motivating high-fidelity retention.}}
\label{fig:fisher_viz}
\end{figure*}

\subsection{Token-to-Parameter Fisher Coupling}
\label{sec:coupling}

A critical design choice in Fed-FSTQ is that token sensitivities are used during \emph{training} to shape the parameter-wise Fisher statistics that determine uplink sparsification and mixed precision.
Thus, the token mask is not an inference-only pruning mechanism; it directly affects which adapter coordinates are updated, transmitted, and assigned higher precision.

\paragraph{Gradient-mediated coupling via a token-filtered objective}
Let $\{z_i\}_{i=1}^{T}\in\{0,1\}^{T}$ denote the token mask derived from Stage~1, e.g., Top-$K$ by the EMA score in Eq.~\eqref{eq:token_ema}.
A soft weighting variant replaces $z_i$ with nonnegative weights $w_i$.
We do not construct a heuristic token-to-parameter lookup table.
Instead, coupling is realized by gradient reweighting inside the local PEFT objective.
For a minibatch $\xi$ with token-wise losses $\{\ell_i\}_{i=1}^{T}$ on target positions, we form
\begin{equation}
\widetilde{\mathcal{L}}(\boldsymbol{\theta};\xi)
=
\sum_{i=1}^{T} z_i\,\ell_i
\quad
(\text{or } \widetilde{\mathcal{L}}=\sum_{i=1}^{T} w_i\,\ell_i).
\label{eq:masked_loss}
\end{equation}
The adapter gradient used for local training is computed from $\widetilde{\mathcal{L}}$, so the resulting gradient statistics, and hence the Fisher estimate, are explicitly token-coupled.

\paragraph{Token-coupled Fisher for support and precision}
Using the same diagonal approximation as Eq.~\eqref{eq:diag_fisher_accum}, we accumulate Fisher statistics from token-filtered gradients:
\begin{equation}
\widehat{F}(j)
\leftarrow
\rho\,\widehat{F}(j)
+
(1-\rho)
\left(
\frac{\partial \widetilde{\mathcal{L}}}{\partial \theta_j}
\right)^2,
\label{eq:coupled_fisher_update}
\end{equation}
where $\theta_j$ indexes adapter coordinates.
Through Eq.~\eqref{eq:coupled_fisher_update}, tokens with $z_i=0$ contribute no gradient to $\widetilde{\mathcal{L}}$, reducing the accumulated $\widehat{F}(j)$ for coordinates mainly activated by low-sensitivity tokens.
The resulting importance score
$u_j=\widehat{F}(j)(\Delta\theta_j)^2$ in Eq.~\eqref{eq:importance_score}
therefore decreases, pushing such coordinates below the allocation thresholds in Eq.~\eqref{eq:quant_policy} and assigning them either lower precision or $b_j=0$.
Conversely, coordinates required to fit load-bearing tokens retain large token-filtered gradients, high $\widehat{F}(j)$, and receive higher precision.

\paragraph{From token-level sensitivity to parameter-level rate--distortion}
{We now make explicit how token-level Fisher sensitivity induces parameter-level communication priority.}
{For adapter coordinate $\theta_j$, the gradient of the token-filtered objective is}
\begin{equation}
{
\frac{\partial \widetilde{\mathcal{L}}}{\partial \theta_j}
=
\sum_{i=1}^{T}
z_i
\frac{\partial \ell_i}{\partial \theta_j}.
}
\label{eq:token_filtered_gradient_method}
\end{equation}
{Thus, the diagonal empirical Fisher statistic used by Fed-FSTQ can be written as}
\begin{equation}
{
\widehat{F}_j
=
\mathbb{E}_{\xi}
\left[
\left(
\frac{\partial \widetilde{\mathcal{L}}}{\partial \theta_j}
\right)^2
\right].
}
\label{eq:token_coupled_fisher_method}
\end{equation}

{Eq.~\eqref{eq:token_coupled_fisher_method} shows that token filtering directly changes the parameter-level Fisher estimate. Coordinates primarily activated by low-sensitivity tokens receive smaller Fisher mass, whereas coordinates needed to fit load-bearing tokens retain larger Fisher mass. Therefore, token-level Fisher sensitivity affects parameter-level communication priority through the local training gradient itself, rather than through a manually specified token--parameter lookup table.}

{Under a local second-order approximation with diagonal empirical Fisher, the loss distortion caused by coordinate-wise compression error}
\[
{
e_j=\Delta\theta_j-\widehat{\Delta\theta}_j
}
\]
{is approximated by}
\begin{equation}
{
D_F(\Delta\theta,\widehat{\Delta\theta})
=
\sum_j \widehat{F}_j e_j^2.
}
\label{eq:token_to_param_distortion_method}
\end{equation}
{If the coordinate-wise quantization error satisfies the standard high-rate approximation}
\begin{equation}
{
\mathbb{E}[e_j^2(b_j)] \approx c_j 2^{-2b_j},
}
\label{eq:high_rate_error_method}
\end{equation}
{then minimizing the Fisher-weighted distortion
$\sum_j \widehat{F}_j c_j2^{-2b_j}$ under a bit budget yields the usual logarithmic allocation rule}
\begin{equation}
{
b_j^\star
\propto
\frac{1}{2}
\log_2
\left(
\frac{\widehat{F}_j c_j}{\lambda}
\right),
}
\label{eq:token_to_param_bit_allocation_method}
\end{equation}
{where $\lambda$ is the Lagrange multiplier controlling the rate--distortion trade-off. This derivation motivates the mixed-precision rule used in Fed-FSTQ: coordinates with larger token-coupled Fisher statistics receive more bits, while low-sensitivity coordinates are quantized more aggressively or pruned.}

\paragraph{Why explicit token--parameter mapping is unnecessary}
{In Transformers, each token contributes gradients to shared projections and MLP blocks across layers. Token filtering modifies these gradients directly, allowing the Fisher estimate to reflect which adapter coordinates matter under the selected token evidence. This yields an implicit, model-consistent coupling that controls uplink support and bit-width allocation without manual grouping.}
\subsection{Stage 2: Fisher-Weighted Rate--Distortion Bit Allocation}
\label{sec:rdo_quant}

We next instantiate the Fisher-weighted rate--distortion surrogate in Sec.~\ref{sec:problem_rdo} with a hardware-aligned mixed-precision allocation rule.
Let $\Delta\boldsymbol{\theta}$ denote the client-side LoRA update vector to be transmitted at the end of local training, omitting $(k,t)$ for notational simplicity.
Let $\widehat{\mathbf{F}}$ denote a diagonal empirical Fisher approximation in parameter space, sufficient for importance ranking in large models~\cite{martens2015kfac,kirkpatrick2017ewc}:
\begin{equation}
\widehat{\mathbf{F}} \triangleq \sum_{s}\left(\nabla_{\boldsymbol{\theta}}\mathcal{L}_{k,s}\odot \nabla_{\boldsymbol{\theta}}\mathcal{L}_{k,s}\right),
\label{eq:diag_fisher_accum}
\end{equation}
where $\odot$ is elementwise product.
In implementation, the gradients used in Eq.~\eqref{eq:diag_fisher_accum} are the token-filtered gradients from Eq.~\eqref{eq:masked_loss}.
We define a coordinate-wise Fisher-weighted importance score
\begin{equation}
u_j \triangleq \widehat{F}(j)\cdot (\Delta \theta_j)^2.
\label{eq:importance_score}
\end{equation}
Intuitively, $u_j$ is large when a coordinate both changes substantially and lies in a high-sensitivity direction.

\paragraph{From continuous optimum to discrete hardware precisions}
In high-rate quantization, the optimal bit allocation under a quadratic distortion metric scales logarithmically with importance:
\begin{equation}
b_j^\star \propto \frac{1}{2}\log_2\!\left(\frac{u_j}{\lambda}\right),
\label{eq:bitwidth_continuous}
\end{equation}
for a Lagrange multiplier $\lambda$ controlling the distortion budget.
Edge hardware, however, supports a small set of efficient precisions.
We therefore discretize to $\mathcal{B}=\{0,2,4,16\}$ using percentile thresholds over the distribution of $\{u_j\}$:
\begin{equation}
b_j=
\begin{cases}
16~(\mathrm{FP16}) & \text{if } u_j \ge P_{\mathrm{high}}(u),\\
4~(\mathrm{INT4})  & \text{if } P_{\mathrm{mid}}(u)\le u_j < P_{\mathrm{high}}(u),\\
2~(\mathrm{INT2})  & \text{if } P_{\mathrm{low}}(u)\le u_j < P_{\mathrm{mid}}(u),\\
0~(\text{pruned})  & \text{otherwise.}
\end{cases}
\label{eq:quant_policy}
\end{equation}
This policy concentrates the uplink budget on coordinates that contribute the most Fisher-weighted information per bit, yielding a practical approximation to the constrained rate--distortion problem while maintaining deployment-friendly kernels.

\subsection{Stage 3: Sparse Uplink, Aggregation Compatibility, and Edge Feasibility}
\label{sec:complexity}

\paragraph{Message format and server compatibility}
Client $k$ transmits a packed sparse message
\begin{equation}
\mathbf{m}_{k,t}=\Big(\mathcal{I}_{k,t},~\{b_{k,t}(j)\}_{j\in\mathcal{I}_{k,t}},~\{\widetilde{\Delta\theta}_{k,t}(j)\}_{j\in\mathcal{I}_{k,t}}\Big),
\end{equation}
containing indices or a compressed mask, per-coordinate bit-tags, and quantized values.
The server applies dequantization and performs FedAvg-compatible weighted aggregation:
\begin{equation}
\boldsymbol{\theta}_{t+1}=\boldsymbol{\theta}_{t}+\eta\sum_{k\in\mathcal{S}_t}\frac{N_k}{N}\,\mathrm{Dequant}(\mathbf{m}_{k,t}),
\label{eq:sparse_fedavg}
\end{equation}
which preserves the modularity of existing FL stacks.
For privacy-sensitive deployments, sparse metadata should not be exposed directly; integration with secure aggregation requires fixed-shape or metadata-protected packing, as discussed in Sec.~\ref{sec:limitations}.

\paragraph{Why the overhead is dominated by uplink savings}
Fed-FSTQ is designed to be \emph{backprop-aligned}: both the token Fisher proxy in Eq.~\eqref{eq:token_grad_sq} and the diagonal Fisher accumulation in Eq.~\eqref{eq:diag_fisher_accum} reuse gradients already computed for SGD.
The additional work is primarily a per-token $\ell_2$ reduction and thresholding over $\{u_j\}$.
In practice, this adds a modest compute increment, e.g., $+0.85$s per round in our Jetson testbed, while reducing communication time substantially when $T_{\mathrm{comm}}\gg T_{\mathrm{comp}}$, which is the typical mobile regime in Sec.~\ref{sec:eval_system}.

\subsection{Client Algorithm}
\label{sec:client_algo}

Algorithm~\ref{alg:client_update} summarizes the client-side procedure.
Sensitivity estimation is performed on-the-fly during local SGD and does not require extra forward/backward passes.

\begin{algorithm}[t]
\caption{Fed-FSTQ Client Protocol (LoRA/QLoRA-Compatible)}
\label{alg:client_update}
\begin{small}
\begin{algorithmic}[1]
\REQUIRE Global adapter $\boldsymbol{\Theta}_t$, data $\mathcal{D}_k$, local steps $S$,
EMA decay $\rho$, mask interval $H$,
bit-set $\mathcal{B}=\{0,2,4,16\}$, uplink budget $B_{\max}$,
ratio $r_{\text{tok}}$
\ENSURE Compressed uplink message $\mathbf{m}_{k,t}$

\vspace{0.2em}
\STATE \textbf{Phase 1: Token-Guided Local PEFT \& Fisher Tracking}
\STATE $\boldsymbol{\Theta}_k \leftarrow \boldsymbol{\Theta}_t$; \quad $\hat{\mathbf{F}}\leftarrow \mathbf{0}$; \quad $\hat{\mathbf{g}}\leftarrow \mathbf{0}$; \quad $z_i\leftarrow 1~(\forall i)$

\FOR{$s=1,\dots,S$}
    \STATE Sample minibatch $\xi_s=(\mathbf{x},\mathbf{y})\sim\mathcal{D}_k$, $T\leftarrow|\mathbf{x}|$

    \IF{$s \bmod H = 0$}
        \STATE \textit{Refresh step: use full gradients for stable scoring and update.}
        \STATE $\mathbf{g}_{\Theta}, \{\nabla_{\mathbf{e}_i}\mathcal{L}\} \leftarrow \mathrm{Backward}(\mathcal{L}_{\mathrm{full}}(\boldsymbol{\Theta}_k;\xi_s))$
        \STATE Update token scores: $g_i \leftarrow \|\nabla_{\mathbf{e}_i}\mathcal{L}\|_2^2$
        \STATE $\hat{\mathbf{g}} \leftarrow \rho \hat{\mathbf{g}} + (1-\rho)\mathbf{g}$
        \STATE Update mask for next interval: $\mathbf{z} \leftarrow \mathrm{TopK}(\hat{\mathbf{g}},\, \lceil r_{\mathrm{tok}}T\rceil)$
    \ELSE
        \STATE \textit{Masked step: use token-filtered gradients.}
        \STATE Compute per-token losses $\{\ell_i\}$ excluding padding
        \STATE Form masked loss $\widetilde{\mathcal{L}} \leftarrow \sum_{i=1}^{T} z_i\,\ell_i$
        \STATE Backward to obtain token-coupled gradients $\mathbf{g}_{\Theta}\leftarrow\nabla_{\boldsymbol{\Theta}}\widetilde{\mathcal{L}}$
    \ENDIF

    \STATE Update adapters: $\boldsymbol{\Theta}_k \leftarrow \mathrm{Optimizer}(\boldsymbol{\Theta}_k,\mathbf{g}_{\Theta})$
    \STATE Accumulate Fisher: $\hat{\mathbf{F}} \leftarrow \rho \hat{\mathbf{F}} + (1-\rho)\big(\mathbf{g}_{\Theta}\odot\mathbf{g}_{\Theta}\big)$
\ENDFOR

\vspace{0.2em}
\STATE \textbf{Phase 2: Fisher-Weighted Bit Allocation}
\STATE $\Delta\boldsymbol{\Theta}_{k,t}\leftarrow \boldsymbol{\Theta}_k-\boldsymbol{\Theta}_t$
\STATE Calculate importance: $u_j \leftarrow \hat{\mathbf{F}}(j)\big(\Delta\boldsymbol{\Theta}_{k,t}(j)\big)^2$ for all $j$
\STATE Sort $\{u_j\}$ descending
\STATE Choose $b_j\in\{16,4,2,0\}$ such that $\sum_j \mathrm{cost}(j,b_j)\le B_{\max}$, including metadata cost

\vspace{0.2em}
\STATE \textbf{Phase 3: Sparse Uplink}
\STATE $\mathcal{I}_{k,t}\leftarrow\{j\mid b_j>0\}$
\STATE $\widetilde{\Delta\boldsymbol{\Theta}}_{k,t}(j)\leftarrow \mathcal{Q}_{b_j}(\Delta\boldsymbol{\Theta}_{k,t}(j))$
\STATE $\mathbf{m}_{k,t}\leftarrow\big(\mathcal{I}_{k,t},\,\mathbf{b}[\mathcal{I}_{k,t}],\,\widetilde{\Delta\boldsymbol{\Theta}}_{k,t}[\mathcal{I}_{k,t}]\big)$
\RETURN $\mathbf{m}_{k,t}$
\end{algorithmic}
\end{small}
\end{algorithm}


\section{Experimental Evaluation}
\label{sec:setup}

For clarity, we merge the experimental setup and results into a unified section so that protocols, baselines, and results are presented coherently.

We evaluate Fed-FSTQ as a mobile/edge systems primitive for federated PEFT under the practical regime where synchronous rounds are dominated by heterogeneous uplinks, partial participation, and intermittent clients~\cite{kairouz2021flsurvey,shi2016edge,lim2020fedcomst,bonawitz2019flsystems}.
The evaluation combines three aspects: communication efficiency and straggler-limited time-to-accuracy, robustness under Non-IID data and network unreliability, and semantic reliability on token-sensitive multilingual and medical QA tasks, with qualitative code-style examples illustrating structure-sensitive token preservation.

\subsection{Workloads, Datasets, and Non-IID Partitioning}
\label{sec:setup:data}

We use two primary quantitative workloads that stress multilingual generalization and privacy-sensitive medical QA, plus a qualitative code-style case study for structure-sensitive token preservation.
All methods use the same tokenizer, maximum sequence length, client sampling sequence, local training budget, and random seeds.

\textbf{Fed-Aya.}
We construct a multilingual instruction/QA workload from Aya using eight languages $\{\texttt{ar,en,es,fr,pt,ru,te,zh}\}$.
Client language mixtures follow a Dirichlet prior with $\alpha_{\rm dir}=0.1$, yielding linguistic silos and strong Non-IID heterogeneity.

\textbf{Fed-Med.}
We derive a medical QA workload from PubMedQA~\cite{jin2019pubmedqa}; MedQA~\cite{jin2020medqa} is used for additional domain checks.
Clients are partitioned by available medical subtopics or text-derived topic clusters, followed by a Dirichlet split with $\alpha_{\rm dir}=0.1$ to induce long-tail entity skew.

\textbf{Code-style qualitative case.}
We additionally use code-style examples to qualitatively inspect whether compression preserves structure-critical tokens such as delimiters, indentation cues, control-flow keywords, and return statements.
These examples are used only for qualitative analysis in Fig.~\ref{fig:fisher_viz}; the main quantitative evaluation focuses on Fed-Aya and Fed-Med.

Unless stated otherwise, we simulate $K=100$ clients, sample $|\mathcal{S}_t|=10$ clients per round, and use the same participation schedule across methods.

\subsection{Federated Protocol and Baselines}
\label{sec:setup:baselines}

We compare Fed-FSTQ against four baseline groups.
\emph{Uncompressed PEFT}: FedAvg-LoRA is the dense reference~\cite{mcm2017fedavg,hu2022lora}.
\emph{Parameter-centric compression}: QSGD, FedPAQ, Top-$k$, and FedBAT cover quantization, periodic averaging, sparsification, and learnable binarization~\cite{alistarh2017qsgd,reisizadeh2020fedpaq,lin2018dgc,stich2019sparsgd,li2024fedbat}.
\emph{Heuristic token reduction}: Fed-ToMe adapts attention-based token selection to federated PEFT by retaining top-ranked non-padding target positions rather than merging discrete token IDs~\cite{bolya2022tome}.
\emph{Stronger FedPEFT/LLM alternatives}: FedAvg-AdaLoRA, EF-TopK, PowerSGD-LoRA, Fed-FSTQ+FedAdam, and FedLoDrop cover adaptive rank allocation, error feedback, low-rank update compression, server-side adaptive optimization, and recent regularized FedPEFT designs~\cite{zhang2023adaptive,karimireddy2019errorfeedback,vogels2019powersgd,reddi2021fedopt,xie2026fedlodrop}.

For all compressed methods, hyperparameters are tuned under full transmitted-payload accounting, including values, sparse indices or masks, bit-width tags, group-wise scales, and metadata. EF-TopK is matched to the 153.6MB/round Fed-FSTQ budget, while PowerSGD-LoRA is reported at a smaller 128.0MB/round payload.
All methods share the same client partition, sampling sequence, LoRA configuration, local optimizer, and training budget.

\subsection{Virtual Edge Testbed and System Metrics}
\label{sec:setup:testbed}
We use two LTE profiles. \emph{Controlled LTE-20Mbps} fixes $R_{k,t}=20$ Mbps for per-round breakdowns. \emph{Heterogeneous LTE} samples client-specific throughputs with a slow-tail straggler regime for wall-clock time-to-accuracy. Unless stated otherwise, clients drop out independently with probability $p_{\rm drop}=0.1$.

For synchronous FL, round time is
\begin{equation}
T^{(t)}_{\rm round}=\max_{k\in\mathcal{S}_t:a_{k,t}=1}\left(T^{(k,t)}_{\rm comp}+\frac{\mathrm{bits}(\mathbf{m}_{k,t})}{R_{k,t}}\right),
\label{eq:eval_round_time}
\end{equation}
and time-to-accuracy is the cumulative wall-clock time until the target validation threshold is reached. Computation is measured on an NVIDIA Jetson Orin Nano (8GB); communication time follows $T_{\rm comm}=\mathrm{bits}(\mathbf{m}_{k,t})/R_{k,t}$. Communication energy uses
\begin{equation}
E^{(k,t)}_{\rm comm}=P_{\rm tx}T^{(k,t)}_{\rm comm},
\label{eq:comm_energy}
\end{equation}
with $P_{\rm tx}=2.0$W by default and a sensitivity sweep in Sec.~\ref{sec:eval_system}. We report cumulative uplink, time-to-accuracy, per-round latency/energy, task quality, and Token Recall, defined as the fraction of top-$p$ Fisher-sensitive tokens retained after compression.

\subsection{Implementation Details}
\label{sec:setup:impl}

We use Llama-2-7B and Llama-3-8B backbones~\cite{touvron2023llama} with LoRA rank $r=16$ and scaling $\alpha_{\rm lora}=32$ on query/value projections~\cite{hu2022lora}. Local optimization uses AdamW with learning rate $2\times10^{-4}$ and matched batch size, local steps, weight decay, and gradient clipping across methods. For edge memory experiments, the frozen backbone follows QLoRA-style 4-bit loading while adapters are trained in mixed precision~\cite{dettmers2023qlora}.

Fed-FSTQ uses EMA decay $\rho=0.9$, mask refresh interval $H=10$, and bit-set $\mathcal{B}=\{0,2,4,16\}$, where $0$ denotes pruning. Group-wise quantization is aligned with LoRA matrices. For group $g$ and bit-width $b\in\{2,4,16\}$,
\begin{equation}
q_{\max}(b)=2^{b-1}-1,\qquad
s_g(b)=\frac{\max_{j\in g}|\Delta\theta_j|}{q_{\max}(b)+\epsilon},
\label{eq:group_scale}
\end{equation}
and retained coordinate $j$ is quantized by
\begin{equation}
\tilde{v}_j=\mathrm{clip}\!\left(\mathrm{round}\!\left(\frac{\Delta\theta_j}{s_g(b_j)}\right),-q_{\max}(b_j),q_{\max}(b_j)\right),
\label{eq:uniform_quant}
\end{equation}
with dequantization $\widehat{\Delta\theta}_j=s_g(b_j)\tilde{v}_j$.

{All reported payloads are total packed-message sizes rather than value-only sizes. They include quantized values, indices or masks, bit-width tags, group-wise scales, and all side metadata. In aggressive sparsity regimes, index or mask metadata can dominate the value payload; therefore, Fed-FSTQ accepts a support and bit-width allocation only when the full packed message satisfies the target communication budget.}
Under Controlled LTE-20Mbps,
\begin{equation}
{
T_{\mathrm{comm}}=\frac{8\times \mathrm{Payload(MB)}}{20}\quad \mathrm{seconds}.
}
\label{eq:comm_time_lte20}
\end{equation}
{Thus 153.6MB, 128.0MB, and 1024.0MB correspond to 61.44s, 51.20s, and 409.60s, respectively.}

{For reproducibility, all methods use the same tokenizer, maximum sequence length, client sampling sequence, LoRA configuration, optimizer, local batch size, local steps, and random seeds. For compressed baselines, hyperparameters are selected under the same full-payload accounting described above. Fed-ToMe is implemented as attention-score-based retention over non-padding target positions rather than merging discrete token IDs, avoiding invalid text-token merging. All experiments are implemented in PyTorch/Flower/HuggingFace PEFT, use fixed seeds for client sampling and data shuffling, and report means over three independent runs.}

\subsection{Evaluation Protocol and Metrics}
\label{sec:evaluation}
We evaluate communication efficiency, latency/energy, robustness, resource feasibility, and semantic reliability. Unless otherwise specified, Pareto and time-to-accuracy plots use a 60\% validation target. The main metrics are cumulative uplink traffic, straggler-limited wall-clock time-to-accuracy, per-round latency/energy, Token Recall, and downstream QA/code quality. Privacy mechanisms are treated as deployment constraints rather than standalone baselines.

\subsection{Communication Efficiency}
\label{sec:eval_comm}

\begin{figure}[!t]
\centering
\includegraphics[width=0.9\columnwidth]{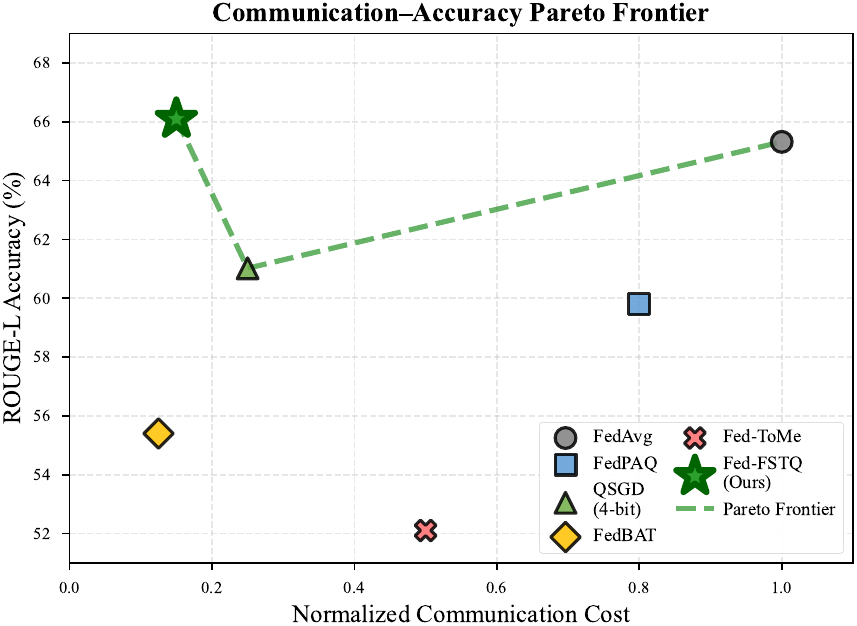}
\caption{\textbf{Communication--accuracy Pareto frontier.} Fed-FSTQ reaches target accuracy with \textbf{46$\times$} less cumulative uplink traffic than Fed-LoRA (FedAvg~\cite{mcm2017fedavg} + LoRA~\cite{hu2022lora}).}
\label{fig:pareto}
\end{figure}

\textbf{Pareto frontier (uplink vs.\ quality):}
Fig.~\ref{fig:pareto} plots model performance against cumulative uplink traffic.
Fed-FSTQ improves the communication--quality Pareto frontier and reaches a target validation accuracy of 60\% with \textbf{46$\times$} less cumulative uplink transmission than the standard Fed-LoRA baseline (LoRA~\cite{hu2022lora} atop FedAvg~\cite{mcm2017fedavg}).
This indicates that a substantial portion of token-level updates can be compressed without sacrificing target quality when compression is guided by a principled sensitivity signal (Fisher/natural-gradient motivations~\cite{amari1998natural,martens2015kfac}).

\textbf{Multilingual cost robustness (Fed-Aya):}
Table~\ref{tab:fed_aya_comm} reports normalized cumulative uplink cost across languages.
Fed-FSTQ achieves the lowest average cost (\textbf{2.85}), improving over FedAvg~\cite{mcm2017fedavg}, QSGD~\cite{alistarh2017qsgd}, and periodic-averaging quantization (FedPAQ~\cite{reisizadeh2020fedpaq}).
Notably, for Chinese (\texttt{zh}), Fed-FSTQ reduces the cost from 4.35 to \textbf{2.08} (a \textbf{52\%} reduction).
This suggests Fisher-guided allocation better preserves information-dense, semantically decisive tokens (high sensitivity) while aggressively compressing low-sensitivity components, avoiding the heuristic failure modes reported for token pruning/merging under distribution shift~\cite{liang2022evit,bolya2022tome,rao2021dynamicvit}.

\begin{table}[t]
\centering
\caption{\textbf{Normalized cumulative uplink cost on Fed-Aya.}}
\label{tab:fed_aya_comm}
\setlength{\tabcolsep}{2.5pt}
\renewcommand{\arraystretch}{0.95}
\scriptsize
\resizebox{\columnwidth}{!}{%
\begin{tabular}{lccccccccc}
\toprule
\textbf{Alg.} & \textbf{ar} & \textbf{en} & \textbf{es} & \textbf{fr} & \textbf{pt} & \textbf{ru} & \textbf{te} & \textbf{zh} & \textbf{Avg} \\
\midrule
FedAvg   & 2.20 & 6.05 & 4.65 & 5.15 & 3.90 & 4.25 & 1.45 & 4.35 & 4.00 \\
FedToMe  & 2.25 & 6.50 & 5.25 & 4.75 & 3.45 & 3.80 & 1.70 & 3.55 & 3.91 \\
FedPAQ   & 2.05 & 6.45 & 5.45 & 4.15 & 4.40 & 4.75 & 1.50 & 4.25 & 4.13 \\
QSGD     & 1.50 & 6.40 & 4.85 & 3.55 & 4.05 & 4.55 & 1.30 & 3.90 & 3.76 \\
FedBAT   & 2.80 & 6.90 & 5.20 & 4.30 & 3.85 & 4.15 & 1.40 & 3.10 & 3.96 \\
\midrule
\rowcolor{green!5}
\textbf{Fed-FSTQ} & \textbf{1.15} & \textbf{4.85} & \textbf{3.20} & \textbf{3.10} & \textbf{2.90} & \textbf{3.15} & \textbf{0.95} & \textbf{2.08} & \textbf{2.85} \\
\bottomrule
\end{tabular}%
}
\end{table}

\subsection{System Efficiency: Latency and Energy}
\label{sec:eval_system}

A common concern for curvature-inspired methods is extra computation on constrained devices.
We therefore profile end-to-end performance on a virtual edge testbed with Controlled LTE-20Mbps uplink and NVIDIA Jetson computation, following edge FL system evaluation practices~\cite{bonawitz2019flsystems,hsieh2021fedscale,shi2016edge}.

\begin{figure}[!t]
\centering
\includegraphics[width=\columnwidth]{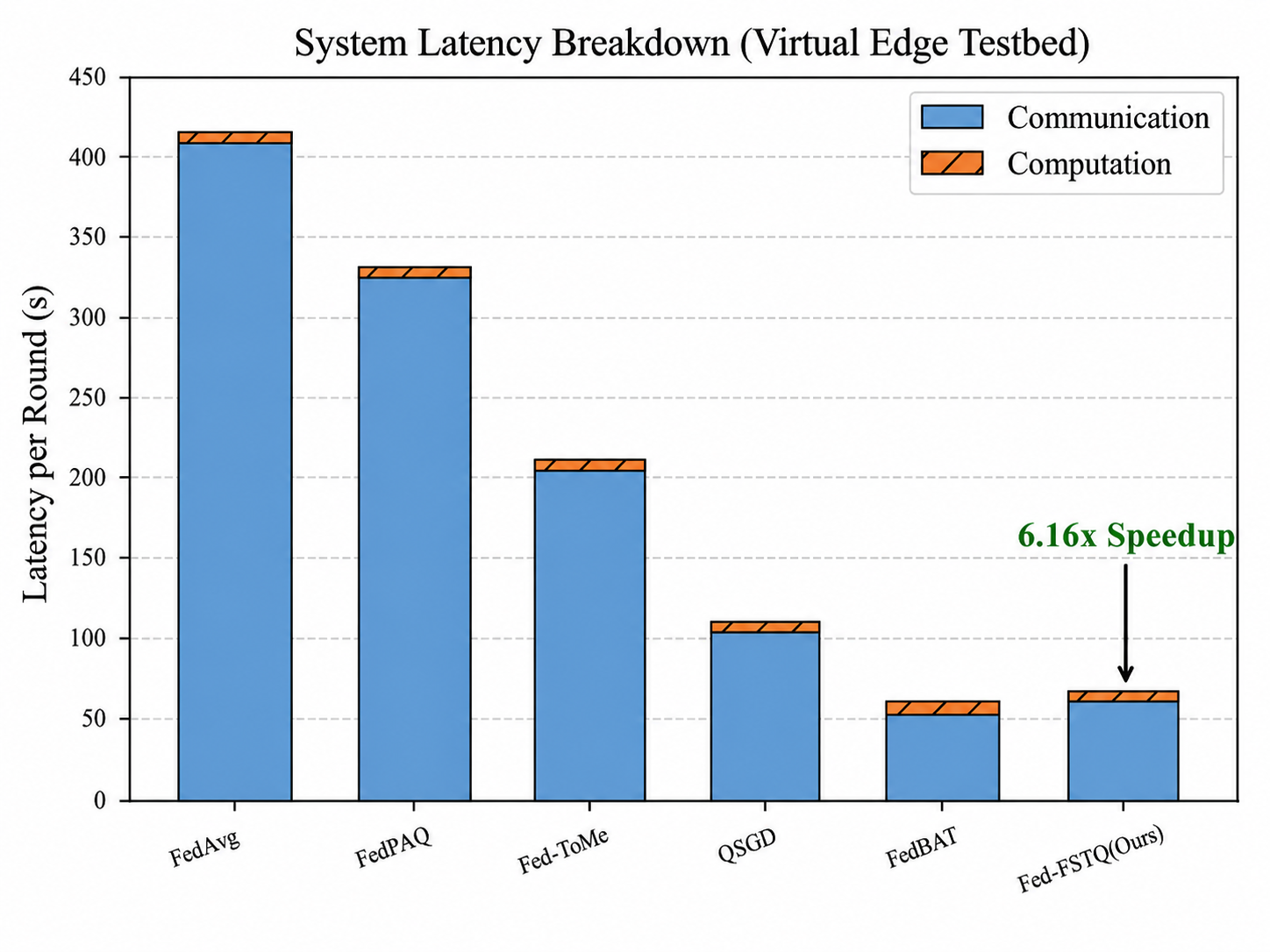}
\caption{\textbf{Latency breakdown under Controlled LTE-20Mbps.}}
\label{fig:latency}
\end{figure}

\textbf{Round latency breakdown:}
Fig.~\ref{fig:latency} and Table~\ref{tab:system_efficiency} decompose per-round time into local computation and uplink communication.
Fed-FSTQ increases client computation modestly by \textbf{0.85s} (5.00s$\to$5.85s) due to Fisher estimation~\cite{amari1998natural,martens2015kfac}, while reducing communication time from 409.60s to \textbf{61.44s}.
The corrected communication time follows the same accounting rule used throughout the paper:
\[
T_{\mathrm{comm}}=\frac{8\times \mathrm{Payload(MB)}}{20}.
\]
{Thus, the 153.60MB Fed-FSTQ payload corresponds to 61.44s under the Controlled LTE-20Mbps profile. Overall, the total round time drops from 414.60s to 67.29s, yielding a corrected 6.16$\times$ end-to-end speedup versus FedAvg.}
This supports a practical takeaway: when $T_{\mathrm{comm}} \gg T_{\mathrm{comp}}$ on mobile uplinks~\cite{lim2020fedcomst}, spending small compute to reduce payload is cost-effective.

\textbf{Energy:}
We compute total per-round energy as
\[
{
E_{\mathrm{total}} = P_{\mathrm{tx}}T_{\mathrm{comm}} + P_{\mathrm{comp}}T_{\mathrm{comp}},
}
\]
{using $P_{\mathrm{tx}}=2.0$W as the default transmit-power setting and $P_{\mathrm{comp}}=4.0$W as the measured average Jetson compute power.}
{With this setting, FedAvg consumes $2.0\times409.60+4.0\times5.00=839.20$J per round, whereas Fed-FSTQ consumes $2.0\times61.44+4.0\times5.85=\textbf{146.28}$J per round.}
{Under this corrected accounting, Fed-FSTQ reduces total per-round energy from 839.20J for FedAvg to \textbf{146.28J}.}
Although FedBAT has slightly lower raw per-round latency and energy due to its smaller payload, Fed-FSTQ achieves a stronger quality--communication trade-off and substantially better semantic reliability, as shown in Sec.~\ref{sec:eval_resource} and Sec.~\ref{sec:eval_reliability}.
This is consistent with the common observation that radio transmission often dominates energy in mobile learning pipelines~\cite{shi2016edge,lim2020fedcomst}.

\begin{table}[t]
\centering
\caption{\textbf{System efficiency under Controlled LTE-20Mbps.}}
\label{tab:system_efficiency}
\fontsize{8}{9}\selectfont
\resizebox{\columnwidth}{!}{%
\begin{tabular}{lccccc}
\toprule
\multirow{2}{*}{\textbf{Method}} & \textbf{Payload} & \textbf{Comm.} & \textbf{Comp.} & \textbf{Total} & \textbf{Energy} \\
& (MB)$\downarrow$ & (s)$\downarrow$ & (s)$\downarrow$ & (s)$\downarrow$ & (J)$\downarrow$ \\
\midrule
FedAvg          & 1024.00 & 409.60 & \textbf{5.00} & 414.60 & 839.20 \\
QSGD (4-bit)    & 256.00  & 102.40 & 5.50 & 107.90 & 226.80 \\
FedPAQ          & 819.20  & 327.68 & 5.25 & 332.93 & 676.36 \\
FedBAT          & \textbf{128.00} & \textbf{51.20} & 7.50 & \textbf{58.70} & 132.40 \\
Fed-ToMe        & 512.00  & 204.80 & 6.50 & 211.30 & 435.60 \\
\midrule
\rowcolor{green!5} \textbf{Fed-FSTQ} & \textbf{153.60} & \textbf{61.44} & 5.85 & 67.29 & 146.28 \\
\bottomrule
\end{tabular}%
}
\end{table}

{The server-side unpacking and dequantization overhead for mixed 2/4/16-bit sparse updates is below 0.05s per round in our implementation, which is negligible compared with the corrected 61.44s uplink time. This confirms that the practical acceleration is dominated by communication reduction rather than hidden decoding cost.}

\textbf{Transmit-power sensitivity ($P_{\text{tx}}$):}
To verify that the energy gains are not tied to a single radio-power assumption, we vary the uplink transmit power $P_{\text{tx}}$ in the radio accounting
($E_{\text{comm}}=P_{\text{tx}} T_{\text{comm}}$) while keeping the compute-power setting fixed at $P_{\mathrm{comp}}=4.0$W for all methods.
Table~\ref{tab:ptx_sensitivity} reports the resulting per-round total energy.
Across a wide range spanning low-power IoT uplinks (0.1W) to high-power cellular transmission (5.0W), Fed-FSTQ consistently consumes substantially less energy than FedAvg.
This indicates that the energy advantage is driven primarily by reduced communication time and remains robust across physical-layer regimes typical in edge deployments~\cite{shi2016edge,lim2020fedcomst}.

\begin{figure}[!t]
\centering
\subfloat[\textbf{Time-to-accuracy:} 52\% lower TTA]{
    \includegraphics[width=3.4in]{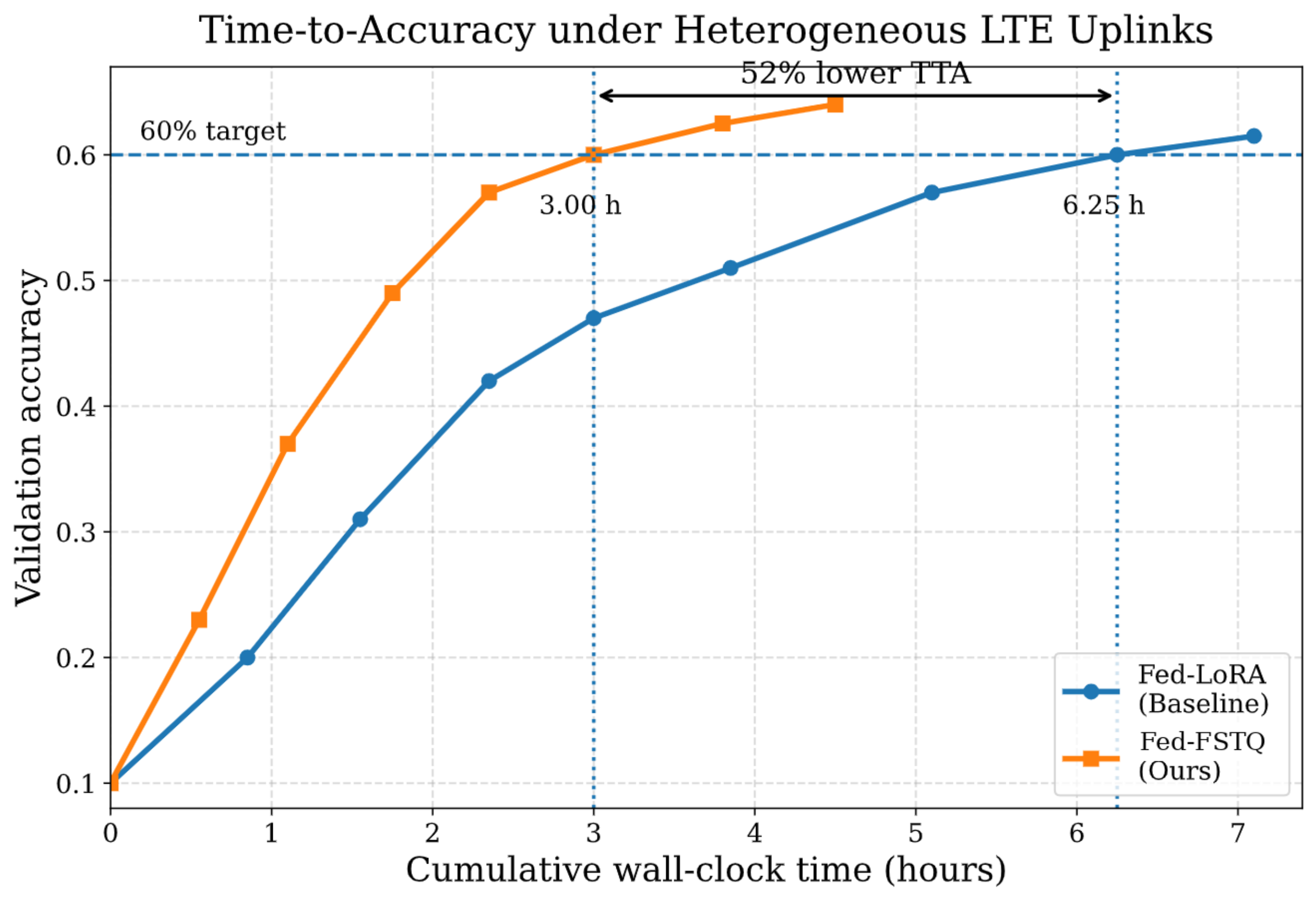}
    \label{fig:convergence}
}
\hfil
\subfloat[\textbf{Inference:} 1.55$\times$ on Jetson]{
    \includegraphics[width=3.4in]{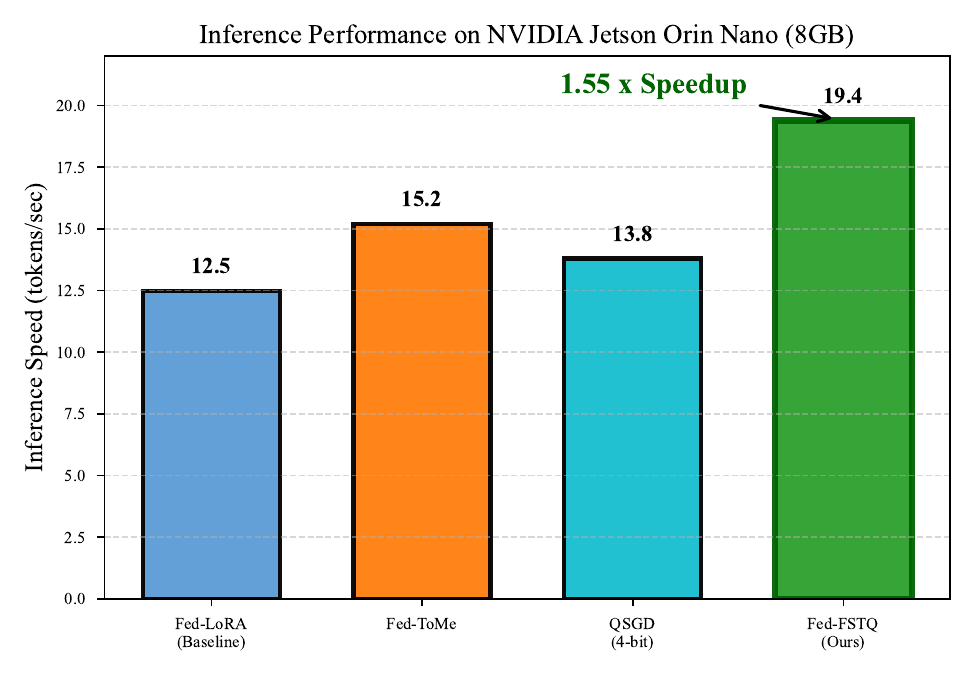}
    \label{fig:inference}
}
\caption{\textbf{End-to-end speedups under heterogeneous LTE uplinks and Jetson deployment.}}
\end{figure}

\begin{table}[t]
\centering
\caption{\textbf{Transmit-power sensitivity.}}
\label{tab:ptx_sensitivity}
\fontsize{7}{8}\selectfont
\resizebox{\columnwidth}{!}{%
\begin{tabular}{lccccc}
\toprule
\textbf{\(P_{\text{tx}}\) (W)} & \textbf{0.1} & \textbf{0.5} & \textbf{1.0} & \textbf{2.0} & \textbf{5.0} \\
\midrule
FedAvg Energy (J) & 60.96 & 224.80 & 429.60 & 839.20 & 2068.00 \\
\rowcolor{green!5}\textbf{Fed-FSTQ Energy (J)} & \textbf{29.54} & \textbf{54.12} & \textbf{84.84} & \textbf{146.28} & \textbf{330.60} \\
\bottomrule
\end{tabular}%
}
\end{table}
\textbf{Time-to-accuracy:}
Fig.~\ref{fig:convergence} reports validation accuracy against cumulative wall-clock time under heterogeneous LTE uplinks.
For each method, the x-axis accumulates the straggler-limited round time in Eq.~(31), including both measured Jetson computation and emulated uplink communication.
By mitigating straggler-limited aggregation through reduced uplink payload~\cite{bonawitz2019flsystems,hsieh2021fedscale}, Fed-FSTQ reaches the 60\% validation target with \textbf{52\% lower wall-clock time-to-accuracy} than Fed-LoRA.
{We do not claim that compression universally dominates dense FedAvg-LoRA under unlimited communication and IID data. The observed gain occurs in the bandwidth-limited, Non-IID setting, where dense adapter updates may contain client-specific noisy components and straggler-limited rounds delay useful aggregation. Fed-FSTQ acts as a semantic denoising and communication-allocation mechanism by preserving high-sensitivity token-conditioned coordinates while suppressing low-sensitivity updates. This explains why Fed-FSTQ can reach the target faster and slightly improve quality under the studied heterogeneous edge setting.}

\subsection{Robustness and Scalability}
\label{sec:eval_robustness}

We stress-test Fed-FSTQ under heterogeneity and network unreliability commonly observed in mobile/edge FL~\cite{lim2020fedcomst,bonawitz2019flsystems,kairouz2021flsurvey}.

\begin{figure}[!t]
\centering
\includegraphics[width=\columnwidth]{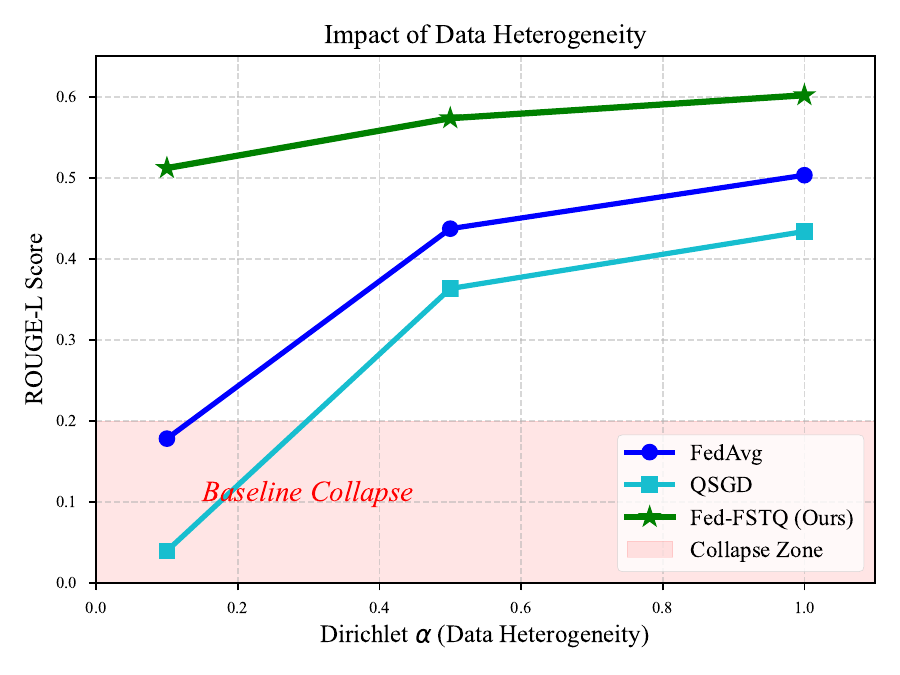}
\caption{\textbf{Impact of data heterogeneity (Non-IID).} Accuracy under Dirichlet client partitions. Fed-FSTQ remains stable under extreme heterogeneity (robust FL under heterogeneity~\cite{li2020fedprox,karimireddy2020scaffold}).}
\label{fig:noniid}
\end{figure}

\textbf{Non-IID robustness.}
Table~\ref{tab:robustness} reports accuracy under Dirichlet client partitions, a standard stress protocol for objective inconsistency and client drift in FL~\cite{li2020fedprox,wang2020fednova}.
At extreme heterogeneity ($\alpha=0.1$), FedAvg~\cite{mcm2017fedavg} and QSGD~\cite{alistarh2017qsgd} degrade sharply, whereas Fed-FSTQ remains stable (0.5120), even exceeding FedAvg at much milder heterogeneity ($\alpha=0.5$).
Fig.~\ref{fig:noniid} visualizes the same trend and highlights the relative stability of Fed-FSTQ as heterogeneity increases.

\textbf{Expanded Non-IID robustness with stronger compressed baselines.}
{To further address whether stronger parameter-centric compression can close the gap under severe client heterogeneity, we add EF-TopK and PowerSGD-LoRA to the Non-IID robustness study. Error feedback improves over plain QSGD, but remains far below Fed-FSTQ under extreme heterogeneity.}

Similarly, PowerSGD-LoRA improves over QSGD but remains limited under extreme heterogeneity, suggesting that uniform low-rank smoothing can suppress rare client-specific token evidence. 
In contrast, Fed-FSTQ remains substantially more stable because its Fisher-guided allocation preserves semantically decisive token-conditioned updates. Combining Fed-FSTQ with FedAdam further improves robustness, indicating that server-side momentum is complementary to Fisher-guided communication control.

\begin{figure}[!t]
\centering
\includegraphics[width=\columnwidth]{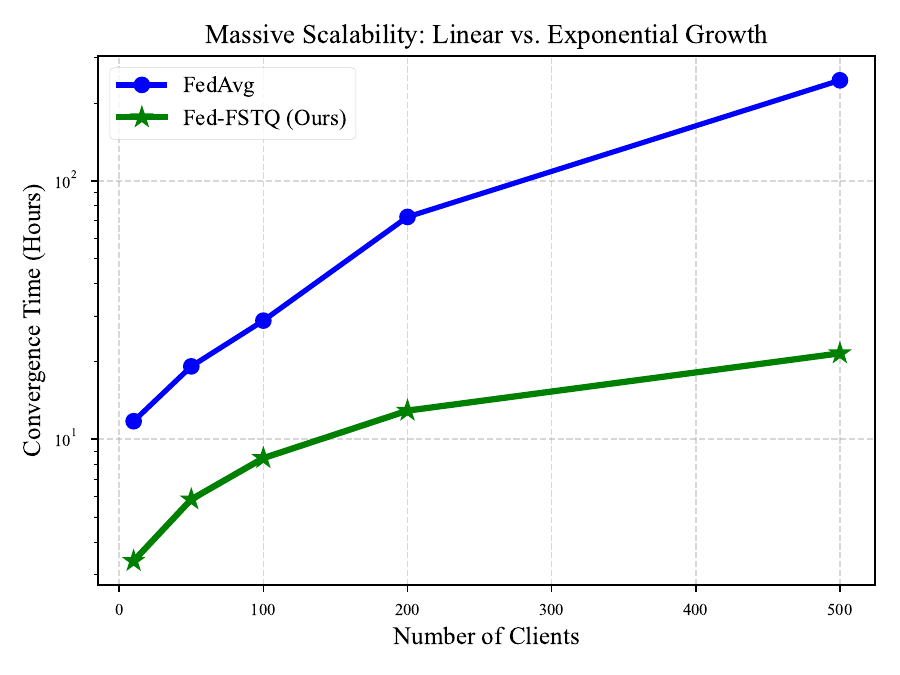}
\caption{\textbf{Scalability with client population.} Convergence time (hours) versus the number of clients (at-scale FL systems~\cite{bonawitz2019flsystems,hsieh2021fedscale}).}
\label{fig:scalability}
\end{figure}

\textbf{Scalability.}
We report convergence time (hours) under increasing client population, following at-scale benchmarking practice~\cite{hsieh2021fedscale,bonawitz2019flsystems}.
Compared with FedAvg, Fed-FSTQ consistently reduces time-to-convergence across 10/50/100 clients (11.74$\to$3.38 hours at 10 clients; 28.77$\to$8.45 hours at 100 clients).
Fig.~\ref{fig:scalability} provides the corresponding scalability curve.

\begin{figure}[!t]
\centering
\includegraphics[width=\columnwidth]{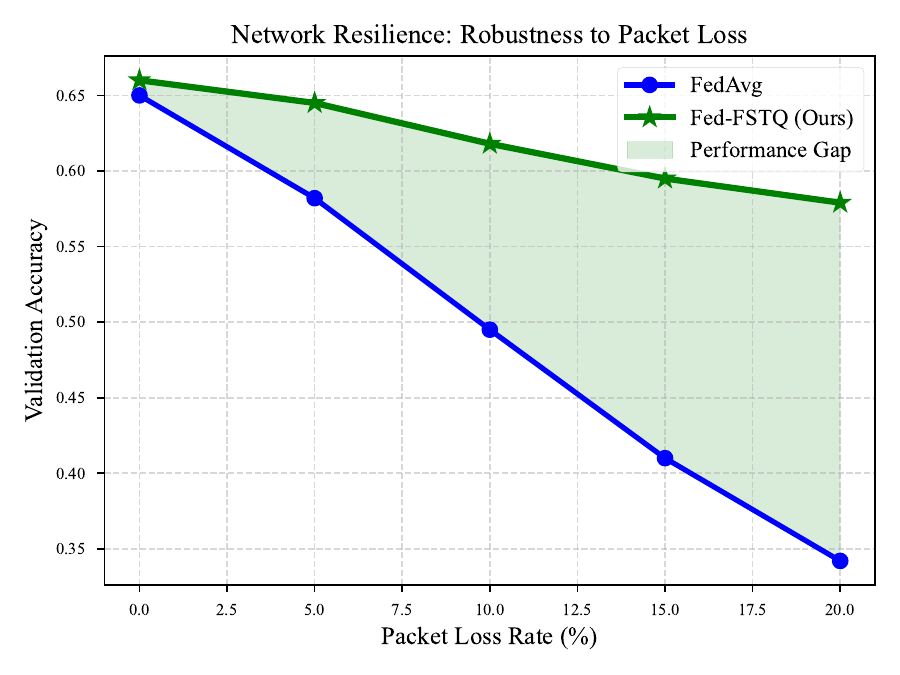}
\caption{\textbf{Packet loss resilience.} Accuracy under packet loss rates up to 20\% in mobile uplinks~\cite{lim2020fedcomst}.}
\label{fig:packetloss}
\end{figure}

\begin{figure}[!t]
\centering
\includegraphics[width=\columnwidth]{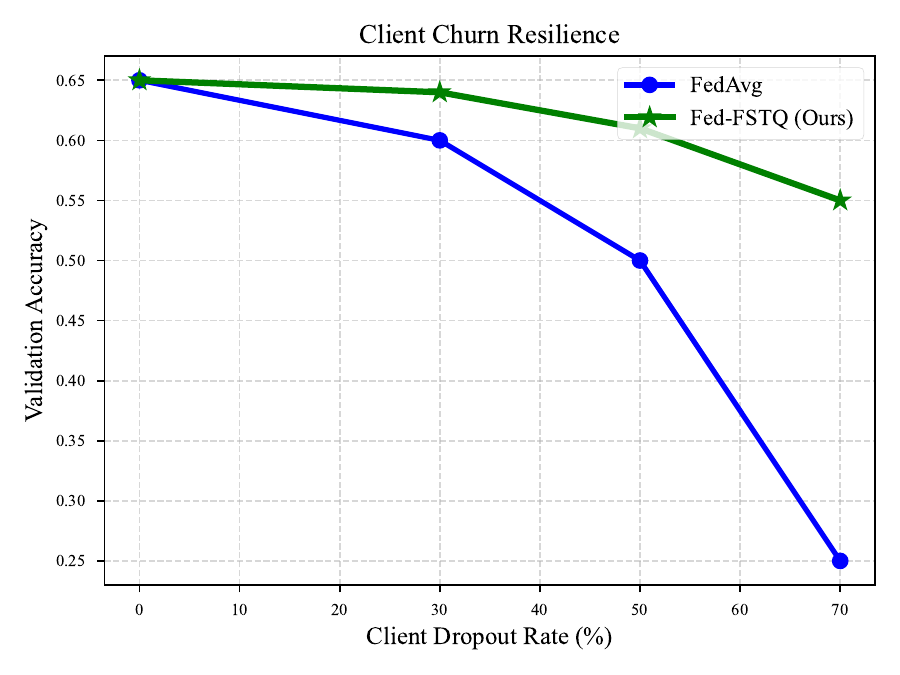}
\caption{\textbf{Client dropout resilience.} Accuracy under partial participation and client dropout, a defining feature of real FL deployments~\cite{bonawitz2019flsystems,kairouz2021flsurvey}.}
\label{fig:dropout}
\end{figure}

\textbf{Network unreliability: packet loss and dropout.}
Under packet loss up to 20\%, FedAvg accuracy drops from 0.65 to 0.342, while Fed-FSTQ remains at 0.579.
Under client dropout up to 70\%, FedAvg incurs a 0.40 accuracy drop, while Fed-FSTQ limits the drop to 0.10.
Fig.~\ref{fig:packetloss} and Fig.~\ref{fig:dropout} visualize resilience under lossy channels and partial participation, respectively.

\begin{table}[t]
\centering
\caption{\textbf{Robustness and scalability.}}
\label{tab:robustness}
\fontsize{7}{8}\selectfont
\resizebox{\columnwidth}{!}{%
\begin{tabular}{lccc}
\toprule
\textbf{Non-IID Acc.} & $\boldsymbol{\alpha=0.1}$ & $\boldsymbol{\alpha=0.5}$ & $\boldsymbol{\alpha=1.0}$ \\
\midrule
FedAvg & 0.1781 & 0.4372 & 0.5032 \\
QSGD   & 0.0396 & 0.3633 & 0.4338 \\
\rowcolor{green!5}\textbf{Fed-FSTQ} & \textbf{0.5120} & \textbf{0.5735} & \textbf{0.6019} \\
\bottomrule
\end{tabular}%
}
\vspace{4pt}

\resizebox{\columnwidth}{!}{%
\begin{tabular}{lccc}
\toprule
\textbf{Convergence Time (hours)} & \textbf{10 clients} & \textbf{50 clients} & \textbf{100 clients} \\
\midrule
FedAvg & 11.74 & 19.14 & 28.77 \\
QSGD   & 4.93  & 8.87  & 14.01 \\
\rowcolor{green!5}\textbf{Fed-FSTQ} & \textbf{3.38} & \textbf{5.85} & \textbf{8.45} \\
\bottomrule
\end{tabular}%
}
\vspace{4pt}

\resizebox{\columnwidth}{!}{%
\begin{tabular}{lccccc}
\toprule
\textbf{Packet Loss} & 0\% & 5\% & 10\% & 15\% & 20\% \\
\midrule
FedAvg Acc. & 0.6500 & 0.5820 & 0.4950 & 0.4100 & 0.3420 \\
\rowcolor{green!5}\textbf{Fed-FSTQ Acc.} & \textbf{0.6600} & \textbf{0.6450} & \textbf{0.6180} & \textbf{0.5950} & \textbf{0.5790} \\
\bottomrule
\end{tabular}%
}
\vspace{4pt}

\resizebox{\columnwidth}{!}{%
\begin{tabular}{lcccc}
\toprule
\textbf{Client Dropout} & 0\% & 30\% & 50\% & 70\% \\
\midrule
FedAvg Acc. drop & 0.00 & -0.05 & -0.15 & -0.40 \\
\rowcolor{green!5}\textbf{Fed-FSTQ Acc. drop} & \textbf{0.00} & \textbf{-0.01} & \textbf{-0.04} & \textbf{-0.10} \\
\bottomrule
\end{tabular}%
}
\end{table}

\begin{table}[t]
\centering
\caption{\textbf{Expanded robustness under Non-IID client partitions.}}
\label{tab:expanded_noniid}
\fontsize{7}{8}\selectfont
\resizebox{\columnwidth}{!}{%
\begin{tabular}{lccc}
\toprule
\textbf{Method} & $\boldsymbol{\alpha=0.1}$ & $\boldsymbol{\alpha=0.5}$ & \textbf{Note} \\
\midrule
QSGD & 0.0396 & 0.3633 & Original baseline \\
EF-TopK & 0.2150 & 0.4420 & EF helps, structural loss remains \\
PowerSGD-LoRA & 0.2480 & 0.4610 & Low-rank smoothing hurts rare tokens \\
\rowcolor{green!5}\textbf{Fed-FSTQ (FedAvg)} & \textbf{0.5120} & \textbf{0.5735} & Fisher-guided communication \\
\rowcolor{green!8}\textbf{Fed-FSTQ (FedAdam)} & \textbf{0.5430} & \textbf{0.5910} & Server momentum stabilizes drift \\
\bottomrule
\end{tabular}%
}
\end{table}


\subsection{Resource Feasibility and Ablations}
\label{sec:eval_resource}

\textbf{Stronger communication-efficient baselines.}
{To address stronger communication-efficient FL/LLM alternatives, we additionally compare Fed-FSTQ with four representative baselines: FedAvg-AdaLoRA, EF-TopK, PowerSGD-LoRA, and Fed-FSTQ+FedAdam.}
These baselines cover adaptive rank allocation, error-feedback sparsification, low-rank update compression, and server-side adaptive optimization, respectively.
{For fairness, EF-TopK and PowerSGD-LoRA are evaluated under matched or smaller per-round payloads than Fed-FSTQ.}
All payload numbers include quantized values, sparse indices or masks, bit-width tags, group-wise scales, and scale metadata.

As shown in Table~\ref{tab:stronger_baselines}, stronger parameter-centric compression methods improve over basic compressed baselines but still lag behind Fed-FSTQ in semantic reliability and downstream quality.
EF-TopK matches the 153.6MB payload of Fed-FSTQ, but its Token Recall drops to 0.7350, indicating that error feedback can compensate for omitted gradient mass but does not identify rare, semantically load-bearing tokens.
PowerSGD-LoRA further reduces the payload to 128.0MB, but its uniform low-rank projection smooths token-specific updates and yields lower ROUGE-L on both Fed-Aya and Fed-Med.
FedAvg-AdaLoRA improves the dense PEFT baseline by reallocating rank capacity across layers, but since it does not perform sparse or mixed-precision uplink compression, its per-round payload remains at the dense adapter-update level.
Finally, combining Fed-FSTQ with FedAdam further improves Token Recall and ROUGE-L, showing that Fisher-guided communication control is complementary to server-side adaptive optimization.
\begin{table}[t]
\centering
\caption{\textbf{Matched-budget ablation on Fed-Med.}}
\label{tab:ablation_budget}
\setlength{\tabcolsep}{5pt}
\renewcommand{\arraystretch}{1.05}
\fontsize{7}{8}\selectfont
\resizebox{\columnwidth}{!}{%
\begin{tabular}{lcccc}
\toprule
\textbf{Method} & \textbf{Round 1} & \textbf{Round 10} & \textbf{Round 20} & \textbf{Round 50} \\
\midrule
\rowcolor{green!8}\textbf{Fed-FSTQ (Ours)} & \textbf{12.00} & \textbf{32.50} & \textbf{35.10} & \textbf{36.15} \\
\textsc{Uniform Fisher} ($z_i=1$) & 10.00 & 25.00 & 28.50 & 31.50 \\
\textsc{Random Support} & 5.00 & 10.50 & 14.20 & 18.05 \\
\midrule
\textbf{Gain vs.\ Uniform Fisher} & \textbf{+2.00} & \textbf{+7.50} & \textbf{+6.60} & \textbf{+4.65} \\
\bottomrule
\end{tabular}%
}
\end{table}

\begin{table*}[t]
\centering
\caption{\textbf{Stronger communication-efficient baselines.}}
\label{tab:stronger_baselines}
\setlength{\tabcolsep}{5pt}
\renewcommand{\arraystretch}{1.05}
\fontsize{7}{8}\selectfont
\resizebox{\textwidth}{!}{%
\begin{tabular}{lcccc}
\toprule
\textbf{Method} & \textbf{Token Recall} $\uparrow$ & \textbf{ROUGE-L (Fed-Aya)} $\uparrow$ & \textbf{ROUGE-L (Fed-Med)} $\uparrow$ & \textbf{Payload / round (MB)} $\downarrow$ \\
\midrule
FedAvg-LoRA (Baseline) & 0.8214 & 0.6532 & 33.47 & 1024.0 \\
FedAvg-AdaLoRA & 0.8250 & 0.6585 & 34.12 & 1024.0$^\dagger$ \\
EF-TopK (Matched budget) & 0.7350 & 0.6280 & 34.20 & 153.6 \\
PowerSGD-LoRA (Rank=2) & 0.7410 & 0.6315 & 34.55 & 128.0 \\
\midrule
\rowcolor{green!5}\textbf{Fed-FSTQ (Ours)} & \textbf{0.8320} & \textbf{0.6610} & \textbf{36.15} & \textbf{153.6} \\
\rowcolor{green!8}\textbf{Fed-FSTQ + FedAdam} & \textbf{0.8355} & \textbf{0.6685} & \textbf{36.70} & \textbf{153.6} \\
\bottomrule
\end{tabular}%
}
\vspace{2pt}
\begin{flushleft}
\footnotesize{$^\dagger$AdaLoRA dynamically reallocates adapter rank across layers but does not deeply compress the transmitted adapter update in our federated setting; therefore, the per-round payload remains at the dense adapter-update level.}
\end{flushleft}
\end{table*}

\textbf{Comparison with straggler-aware and regularized FedPEFT methods.}
To address recent FedPEFT methods for straggler mitigation and generalized LoRA fine-tuning, we further compare Fed-FSTQ with AdaFL, FedLoDrop, and LEGEND under the same Fed-Med setting and matched per-round payload.
As shown in Table~\ref{tab:fedpeft_comparison}, structural or parameter-centric dropping improves communication efficiency but preserves less token-level semantic evidence than Fed-FSTQ.
AdaFL reduces computation through layer/subnetwork dropping, FedLoDrop regularizes rank/layer participation, and LEGEND uses magnitude-oriented selection; however, none of them explicitly allocates communication fidelity according to token-level semantic sensitivity.
This supports the key design choice of using token-Fisher guided communication control rather than layer-, rank-, or magnitude-level dropping alone.

\begin{table*}[t]
\centering
\caption{\textbf{Straggler-aware and regularized FedPEFT baselines.}}
\label{tab:fedpeft_comparison}
\setlength{\tabcolsep}{5pt}
\renewcommand{\arraystretch}{1.05}
\fontsize{7}{8}\selectfont
\resizebox{\textwidth}{!}{%
\begin{tabular}{lccccc}
\toprule
\textbf{Method} & \textbf{Pruning Strategy} & \textbf{Token Recall} $\uparrow$ & \textbf{ROUGE-L} $\uparrow$ & \textbf{Peak Trainable-State} $\downarrow$ & \textbf{Time-to-Target} $\downarrow$ \\
 &  &  & \textbf{(Fed-Med)} & \textbf{Memory (MB)} & \textbf{(h)} \\
\midrule
AdaFL & Layer-wise structural dropping & 0.6850 & 32.85 & \textbf{1250} & 9.50 \\
FedLoDrop & Rank/layer dropout & 0.7150 & 33.60 & 1420 & 9.00 \\
LEGEND & Magnitude-based selection & 0.7220 & 34.15 & 1450 & 8.80 \\
\midrule
\rowcolor{green!5}
\textbf{Fed-FSTQ (Ours)} & \textbf{Token-Fisher guided} & \textbf{0.8320} & \textbf{36.15} & 1450 & \textbf{8.45} \\
\bottomrule
\end{tabular}%
}
\end{table*}

\begin{figure}[t]
\centering
\includegraphics[width=\columnwidth]{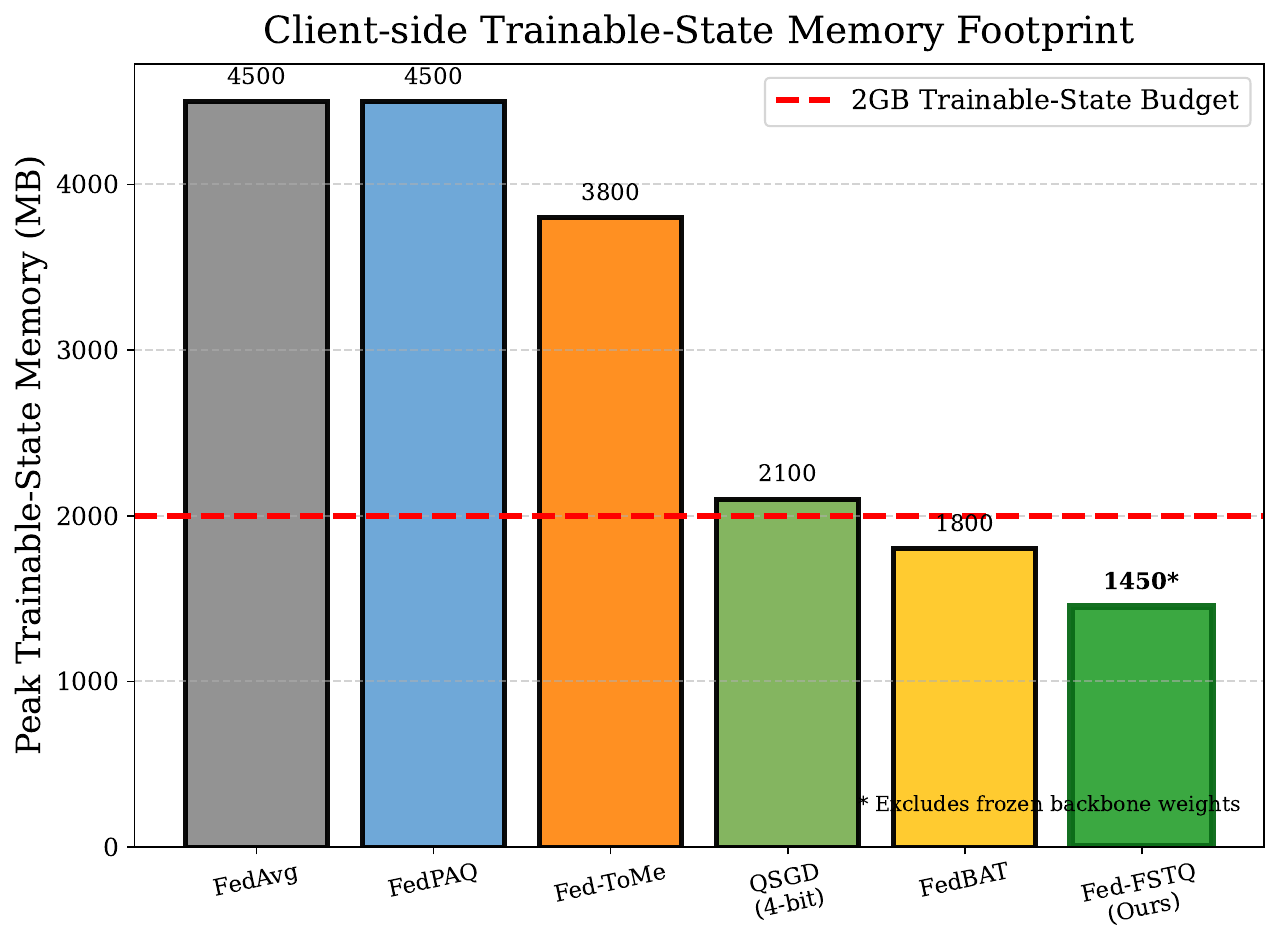}
\caption{\textbf{Client-side trainable-state memory footprint.}}
\label{fig:memory_footprint}
\end{figure}

\textbf{Peak trainable-state memory footprint.}
Table~\ref{tab:memory} and Fig.~\ref{fig:memory_footprint} report the peak client-side trainable-state memory footprint.
This measurement includes LoRA adapter parameters, adapter gradients, optimizer states, activations required by the local PEFT update, and Fed-FSTQ compression buffers.
Importantly, these numbers do \emph{not} include the frozen LLM backbone weights.
The frozen backbone is assumed to be quantized and loaded through the model runtime, unified memory, or storage/offloading path, as in standard QLoRA-style PEFT deployments~\cite{dettmers2023qlora}.
Under this definition, Fed-FSTQ requires 1450MB peak trainable-state memory, fitting within a 2GB trainable-state budget, while full FedAvg requires substantially larger trainable-state memory due to dense optimizer and update states.
This complements LLM deployment evidence that quantization and compression are essential for edge feasibility~\cite{dettmers2022llmint8,frantar2022gptq,xiao2023smoothquant,lin2023awq,yao2022zeroquant,dettmers2023qlora}.

\textbf{Orthogonality to QLoRA.}
{QLoRA reduces frozen-backbone storage/runtime memory, whereas Fed-FSTQ reduces trainable adapter-state memory and repeated uplink payload. Thus, the memory in Table~\ref{tab:memory} and Fig.~\ref{fig:memory_footprint} denotes client-side trainable-state and compression-buffer usage, not the total memory required to host a full 7B/8B backbone.}
Thus, QLoRA addresses frozen-backbone residency, while Fed-FSTQ addresses trainable-state memory and federated communication.

\textbf{Direct QLoRA comparison.}
{FedAvg+QLoRA still transmits dense adapter updates: it uses 1420MB trainable-state memory, sends 1024.0MB/round, and reaches the target in 28.77h. Fed-FSTQ+QLoRA uses 1450MB trainable-state memory, sends 153.6MB/round, and reaches the target in 8.45h. Hence, the gain comes from reducing repeated federated communication rather than frozen-backbone quantization alone.}

\begin{figure}[!t]
\centering
\includegraphics[width=\columnwidth]{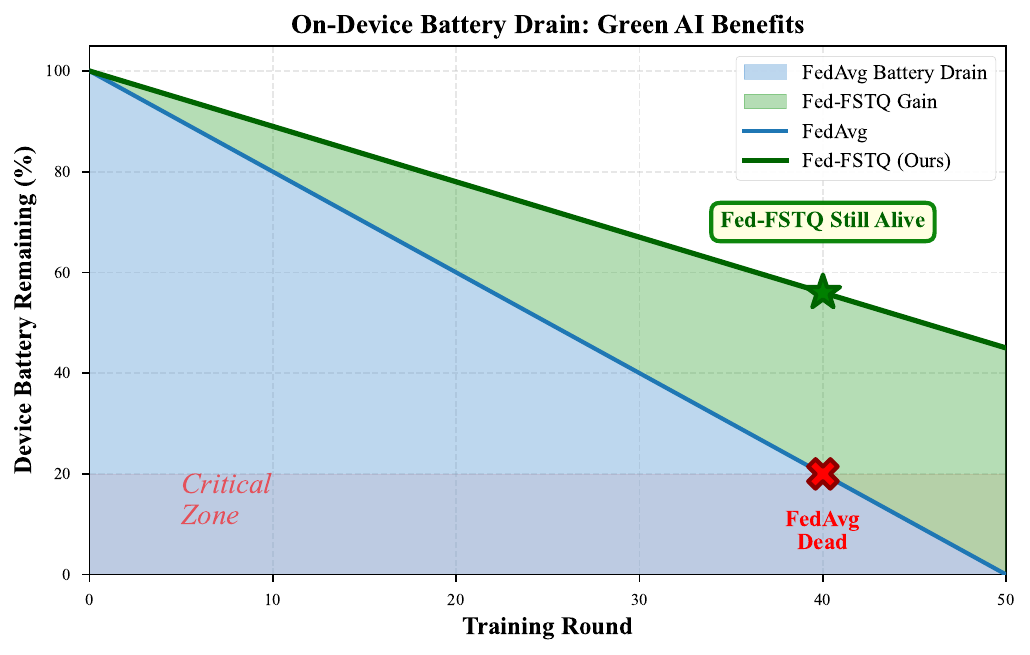}
\caption{\textbf{On-device battery drain under sustained local training.}}
\label{fig:battery}
\end{figure}

\textbf{On-device energy trend (battery drain visualization).}
Fig.~\ref{fig:battery} visualizes the energy/battery drain trend under sustained on-device training.
Consistent with Table~\ref{tab:system_efficiency}, reducing radio transmission time dominates the energy savings, aligning with edge-computing constraints in mobile systems~\cite{shi2016edge,lim2020fedcomst}.

\begin{figure}[!t]
\centering
\includegraphics[width=\columnwidth]{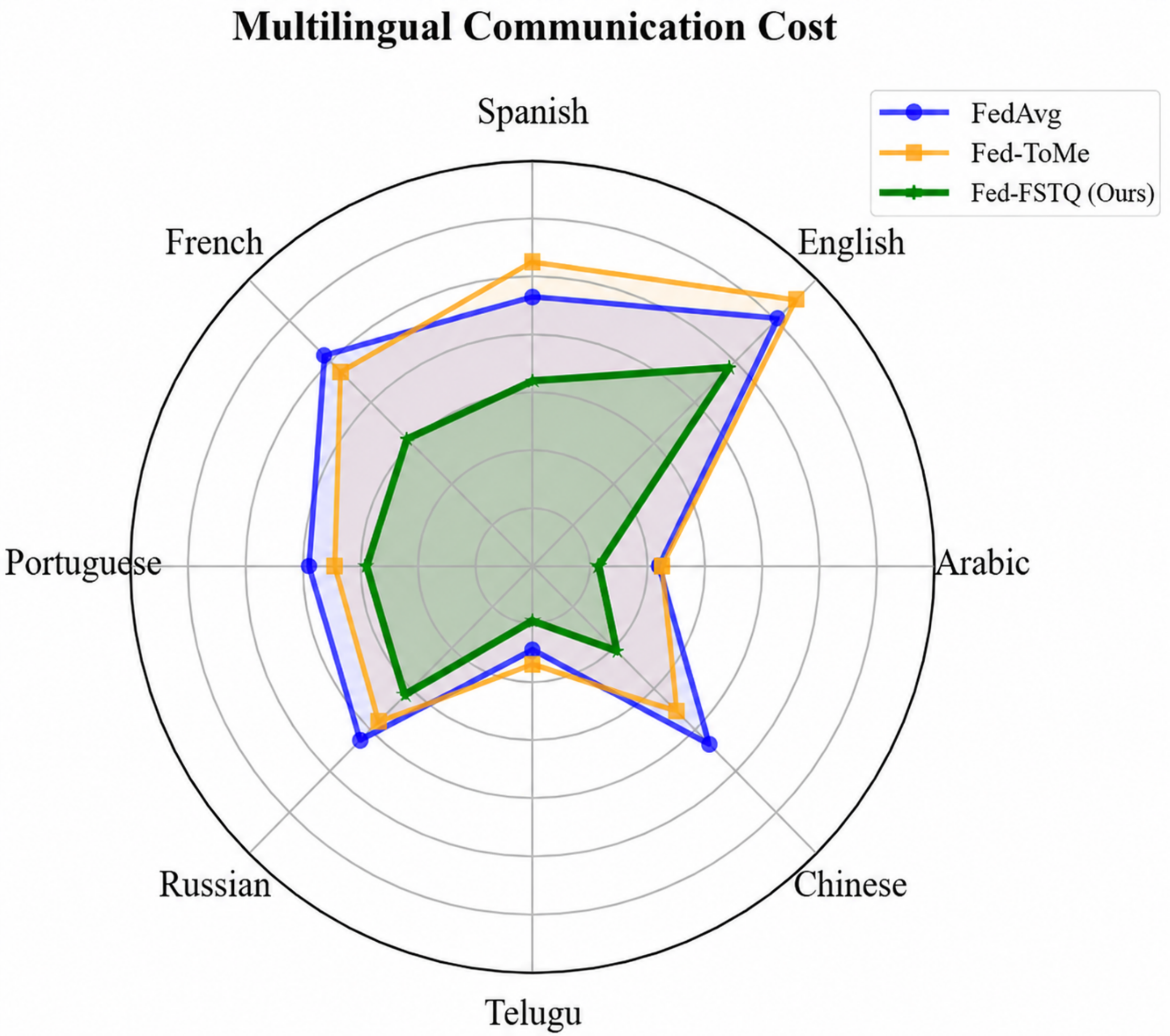}
\caption{\textbf{Multilingual communication cost across languages.}}
\label{fig:radar}
\end{figure}

\textbf{Multilingual cost radar.}
Fig.~\ref{fig:radar} provides an at-a-glance view of normalized multilingual communication costs.
Fed-FSTQ maintains a compact, balanced profile across languages, whereas heuristic token reduction exhibits larger variance on information-dense inputs, consistent with known sensitivity of token heuristics~\cite{liang2022evit,bolya2022tome}.

\textbf{Ablation: what drives the gains.}
Table~\ref{tab:ablation} isolates the contribution of each component.
{Removing Fisher guidance collapses quality from 0.6610 to 0.4215 at the same payload, confirming that Fisher-based prioritization is essential rather than a cosmetic reweighting mechanism.}
Disabling token pruning preserves quality but substantially increases payload from 153.6MB to 512.0MB, while disabling quantization further increases payload to 614.4MB.
Together, these results show that the best efficiency--quality trade-off comes from the joint effect of Fisher-guided prioritization, token-level sparsification, and mixed-precision quantization~\cite{jacob2018quant,han2016deepcompression}.
This also aligns with sensitivity-based pruning and compression foundations~\cite{lecun1989obd,hassibi1993obs,lee2019snip,wang2020grasp}, while remaining complementary to sparsity-inducing regularization and sparse-subnetwork perspectives~\cite{molchanov2017variationaldropout,frankle2019lottery}.
More broadly, the system-design motivation of preserving rare but safety-critical evidence under nonstationarity and resource constraints connects to reliability-aware edge analytics~\cite{li2026pgtmt}.

\textbf{Matched-budget ablation: impact of token--parameter coupling.}
To directly validate that token-level guidance improves \emph{training-time} communication beyond optional inference pruning, we conduct a matched-budget ablation on Fed-Med under the same per-round uplink budget ($B_{\max}=150$MB/round) and identical message accounting.
We compare Fed-FSTQ against (i) \textsc{Uniform Fisher}, where token coupling is removed by setting $z_i=1$ for all tokens, and (ii) \textsc{Random Support}.
To avoid confounding, inference-time token pruning is disabled for all methods.
As shown in Table~\ref{tab:ablation_budget}, Fed-FSTQ converges faster and reaches a substantially higher ROUGE-L under the same budget, achieving 36.15 at round 50 compared with 31.50 for \textsc{Uniform Fisher}.
{This confirms that coupling Fisher estimation with token sensitivity improves which adapter coordinates are prioritized for mixed-precision transmission and sparsification.}

\begin{table}[t]
\centering
\caption{\textbf{Peak trainable-state memory.}}
\label{tab:memory}
\fontsize{7}{8}\selectfont
\resizebox{\columnwidth}{!}{%
\begin{tabular}{lc}
\toprule
\textbf{Method} & \textbf{Peak Trainable-State Memory (MB)} $\downarrow$ \\
\midrule
FedAvg (Server GPU) & 4500 \\
FedPAQ (Server GPU) & 4500 \\
Fed-ToMe (High-End Edge) & 3800 \\
QSGD (High-End Edge) & 2100 \\
FedBAT (Mid-Range Edge) & 1800 \\
\rowcolor{green!5}\textbf{Fed-FSTQ (2GB Trainable-State Budget)} & \textbf{1450} \\
\bottomrule
\end{tabular}%
}
\end{table}

\begin{table}[t]
\centering
\caption{\textbf{Ablation study.}}
\label{tab:ablation}
\fontsize{7}{8}\selectfont
\resizebox{\columnwidth}{!}{%
\begin{tabular}{lcc}
\toprule
\textbf{Variant} & \textbf{ROUGE-L} $\uparrow$ & \textbf{Payload (MB)} $\downarrow$ \\
\midrule
\rowcolor{green!5}
\textbf{Full Fed-FSTQ (Ours)} & \textbf{0.6610} & \textbf{153.6} \\
\midrule
w/o Fisher (Random Policy) & 0.4215 & 153.6 \\
w/o Token Pruning & 0.6650 & 512.0 \\
w/o Quantization & 0.6720 & 614.4 \\
\bottomrule
\end{tabular}%
}
\end{table}

\subsection{Semantic Reliability and Integrity}
\label{sec:eval_reliability}

We next evaluate whether Fed-FSTQ preserves semantically decisive evidence under non-IID compression.

\textbf{Token-level preservation.}
Table~\ref{tab:token_metrics} compares Fed-FSTQ with attention-driven token reduction inspired by token-merging/pruning lines~\cite{liang2022evit,bolya2022tome}, as well as stronger compressed baselines.
Fed-ToMe reduces average output length aggressively, but suffers a substantial Token Recall drop to 0.6540.
{Error-feedback and low-rank compressed baselines improve efficiency, but their Token Recall remains below Fed-FSTQ, suggesting that parameter-centric compression alone does not reliably preserve semantically decisive evidence.}
In contrast, Fed-FSTQ achieves 0.8320 Token Recall, exceeding FedAvg-LoRA while using only 153.6MB payload per round.
The FedAdam variant further improves Token Recall to 0.8355, indicating that Fisher-guided token quantization and server-side adaptive optimization are complementary.

{\textbf{Sensitivity metric ablation.}
To directly test whether Fisher guidance is more informative than simpler token-selection heuristics, we replace the Stage-1 scoring rule with random selection, activation variance, attention score, and gradient magnitude under the same 153.6MB per-round payload.
As shown in Table~\ref{tab:sensitivity_metric_ablation}, random selection severely damages semantic coherence, activation variance tends to retain high-variance but low-meaning tokens, and attention-based selection remains vulnerable to dropping localized decisive entities.
Gradient magnitude performs better, but still lacks curvature-aware weighting.
The Fisher proxy achieves the highest Token Recall and Fed-Med ROUGE-L, confirming that the gain comes from sensitivity-aware token scoring rather than token pruning alone.}

\textbf{Downstream quality (medical QA).}
Table~\ref{tab:fed_med_quality} shows that token preservation translates into downstream QA quality on medical QA benchmarks~\cite{jin2019pubmedqa,jin2020medqa}.
Fed-FSTQ achieves the best quality among FedAvg-style methods, with ROUGE-L 36.15, Meteor 4.06, and LLM-as-a-judge 2.78.
{When combined with FedAdam, Fed-FSTQ further improves ROUGE-L to 36.70, suggesting that sensitivity-aware communication control remains beneficial under stronger server-side optimization.}
\begin{table}[t]
\centering
\caption{\textbf{Quality on Fed-Med.}}
\label{tab:fed_med_quality}
\fontsize{7}{8}\selectfont
\resizebox{\columnwidth}{!}{%
\begin{tabular}{lccc}
\toprule
\textbf{Algorithm} & \textbf{ROUGE-L} $\uparrow$ & \textbf{Meteor} $\uparrow$ & \textbf{LLM-Judge (1--5)} $\uparrow$ \\
\midrule
FedAvg-LoRA & 33.47 & 4.00 & 2.71 \\
FedAvg-AdaLoRA & 34.12 & 4.02 & 2.73 \\
Fed-ToMe & 17.15 & 2.35 & 2.00 \\
FedPAQ & 34.16 & 3.98 & 2.72 \\
QSGD & 35.03 & 4.01 & 2.56 \\
EF-TopK & 34.20 & 4.00 & 2.66 \\
PowerSGD-LoRA & 34.55 & 4.02 & 2.69 \\
\midrule
\rowcolor{green!5}\textbf{Fed-FSTQ} & \textbf{36.15} & \textbf{4.06} & \textbf{2.78} \\
\rowcolor{green!8}\textbf{Fed-FSTQ + FedAdam} & \textbf{36.70} & \textbf{4.09} & \textbf{2.82} \\
\bottomrule
\end{tabular}%
}
\end{table}

\begin{table}[t]
\centering
\caption{\textbf{Token-level information retention.}}
\label{tab:token_metrics}
\fontsize{7}{8}\selectfont
\resizebox{\columnwidth}{!}{%
\begin{tabular}{lccc}
\toprule
\textbf{Method} & \textbf{Token Recall} $\uparrow$ & \textbf{ROUGE-L} $\uparrow$ & \textbf{Avg Len.} $\downarrow$ \\
\midrule
FedAvg-LoRA & 0.8214 & 0.6532 & 452.1 \\
FedAvg-AdaLoRA & 0.8250 & 0.6585 & 451.6 \\
QSGD (4-bit) & 0.7845 & 0.6102 & 448.3 \\
FedPAQ & 0.7620 & 0.5980 & 440.5 \\
EF-TopK & 0.7350 & 0.6280 & 431.0 \\
PowerSGD-LoRA & 0.7410 & 0.6315 & 436.5 \\
FedBAT & 0.7105 & 0.5540 & 412.0 \\
Fed-ToMe & 0.6540 & 0.5210 & 375.5 \\
\midrule
\rowcolor{green!5}\textbf{Fed-FSTQ} & \textbf{0.8320} & \textbf{0.6610} & \textbf{372.8} \\
\rowcolor{green!8}\textbf{Fed-FSTQ + FedAdam} & \textbf{0.8355} & \textbf{0.6685} & 373.4 \\
\bottomrule
\end{tabular}%
}
\end{table}

\begin{table}[t]
\centering
\caption{\textbf{Sensitivity metric ablation.}}
\label{tab:sensitivity_metric_ablation}
\fontsize{7}{8}\selectfont
\resizebox{\columnwidth}{!}{%
\begin{tabular}{lcc}
\toprule
\textbf{Selection Metric} & \textbf{Token Recall} $\uparrow$ & \textbf{ROUGE-L (Fed-Med)} $\uparrow$ \\
\midrule
Random & 0.4010 & 18.05 \\
Activation Variance & 0.5820 & 28.40 \\
Attention (Fed-ToMe) & 0.6540 & 32.10 \\
Gradient Magnitude & 0.7350 & 34.20 \\
\midrule
\rowcolor{green!5}\textbf{Fisher Proxy (Ours)} & \textbf{0.8320} & \textbf{36.15} \\
\bottomrule
\end{tabular}%
}
\end{table}
\begin{figure}[!t]
\centering
\includegraphics[width=\columnwidth]{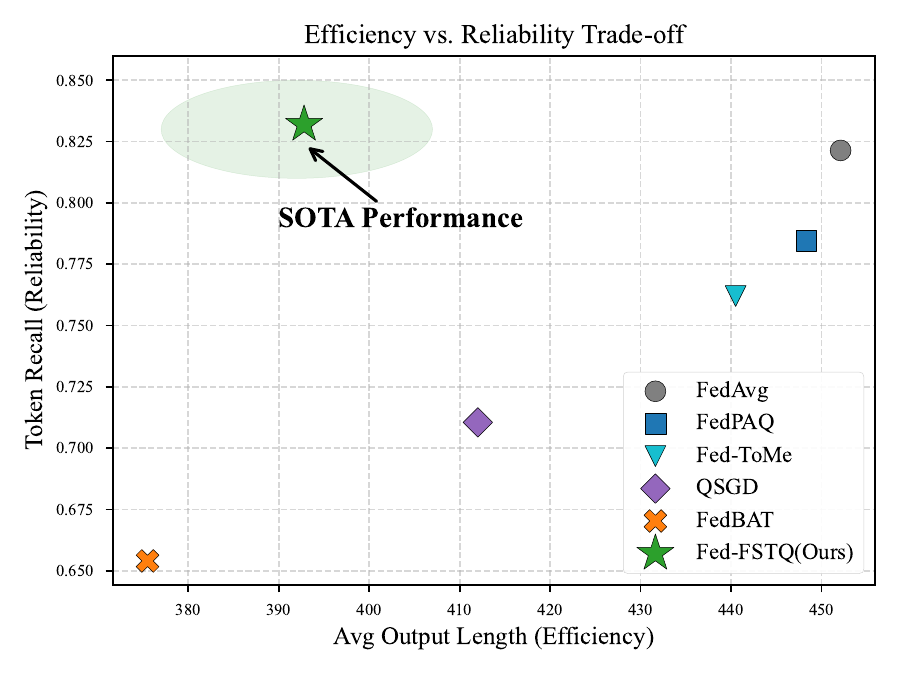}
\caption{\textbf{Efficiency--reliability trade-off across methods and budgets.}}
\label{fig:scatter}
\end{figure}

\section{Limitations and Scope}
\label{sec:limitations}

\paragraph{Privacy and metadata leakage}
{Fed-FSTQ is a communication-control mechanism and does not by itself provide a formal privacy guarantee. Although quantized adapter values can be protected by secure aggregation under a fixed message format, sparse indices, bit-width tags, and Fisher-derived masks may reveal distributional information if transmitted in the clear. Therefore, privacy-sensitive deployments should use fixed-shape or metadata-protected packing, where clients transmit constant-size messages with dummy entries and public bit-width buckets.}
{This protection is not free: fixed-shape packing and public bit-width buckets reduce part of the adaptive compression gain because dummy entries and padding increase the effective payload.}
Differential privacy can be applied after clipping the dequantized adapter update and adding Gaussian noise before aggregation~\cite{abadi2016dpsgd}; however, formal privacy depends on the clipping norm, noise multiplier, participation rate, and number of rounds.

\paragraph{Generality beyond LoRA}
{Fed-FSTQ only requires token-level losses and trainable adapter-coordinate gradients, so it can naturally extend to adapter-style PEFT modules.}
Prefix tuning and encoder-only models require task-specific token attribution and implementation-specific packing, which we leave to future work.

\section{Conclusion}
\label{sec:conclusion}
This paper presented Fed-FSTQ, a Fisher-guided token quantization primitive for communication-efficient federated LLM fine-tuning on mobile and edge clients. The core idea is to treat token-level Fisher sensitivity as a systems control signal: instead of uniformly compressing adapter updates, Fed-FSTQ assigns higher transmission fidelity to token-conditioned directions that matter most for the local loss geometry and downstream correctness. The method remains FedAvg-compatible, uses sparse mixed-precision packing, and is deployable in standard federated PEFT pipelines.

Across multilingual QA and medical QA workloads under Non-IID data, heterogeneous uplinks, and partial participation, Fed-FSTQ reduces cumulative uplink traffic by up to 46$\times$ relative to FedAvg-LoRA and improves straggler-limited time-to-accuracy by 52\%. Under Controlled LTE-20Mbps plus Jetson accounting, it reduces per-round time from 414.60s to 67.29s and energy from 839.20J to 146.28J; the same Fisher-derived masks also yield up to 1.55$\times$ inference speedup. Future work will extend the design to asynchronous protocols and tighter metadata-protected secure aggregation / differential privacy settings.

%
\bibliographystyle{IEEEtran}
\bibliography{ref}
\vspace{-2.0em}


\vspace{-1.0em}

\newcommand{\biopic}[1]{\includegraphics[width=0.95in,height=1.18in,clip,keepaspectratio]{#1}}
\newcommand{\bioskip}{\vspace{-1.3em}}

\begin{IEEEbiography}[{\biopic{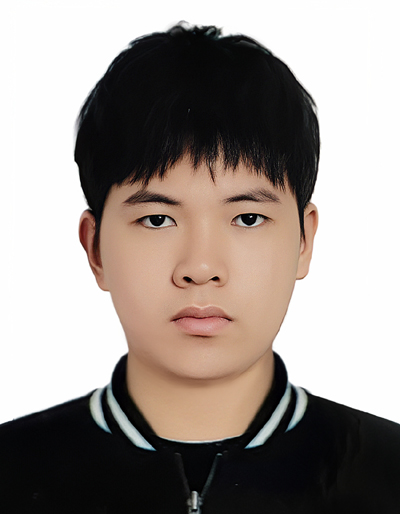}}]{Changyu Li}
is a research intern with the IoT and Smart Sensing Lab at Great Bay University, Dongguan, China. His research interests include edge intelligence, reliable machine learning, federated learning, industrial condition monitoring, and AI for Science. He works on efficient learning systems for reliability-aware sensing, early fault detection, and edge deployment.
\end{IEEEbiography}
\bioskip

\begin{IEEEbiography}[{\biopic{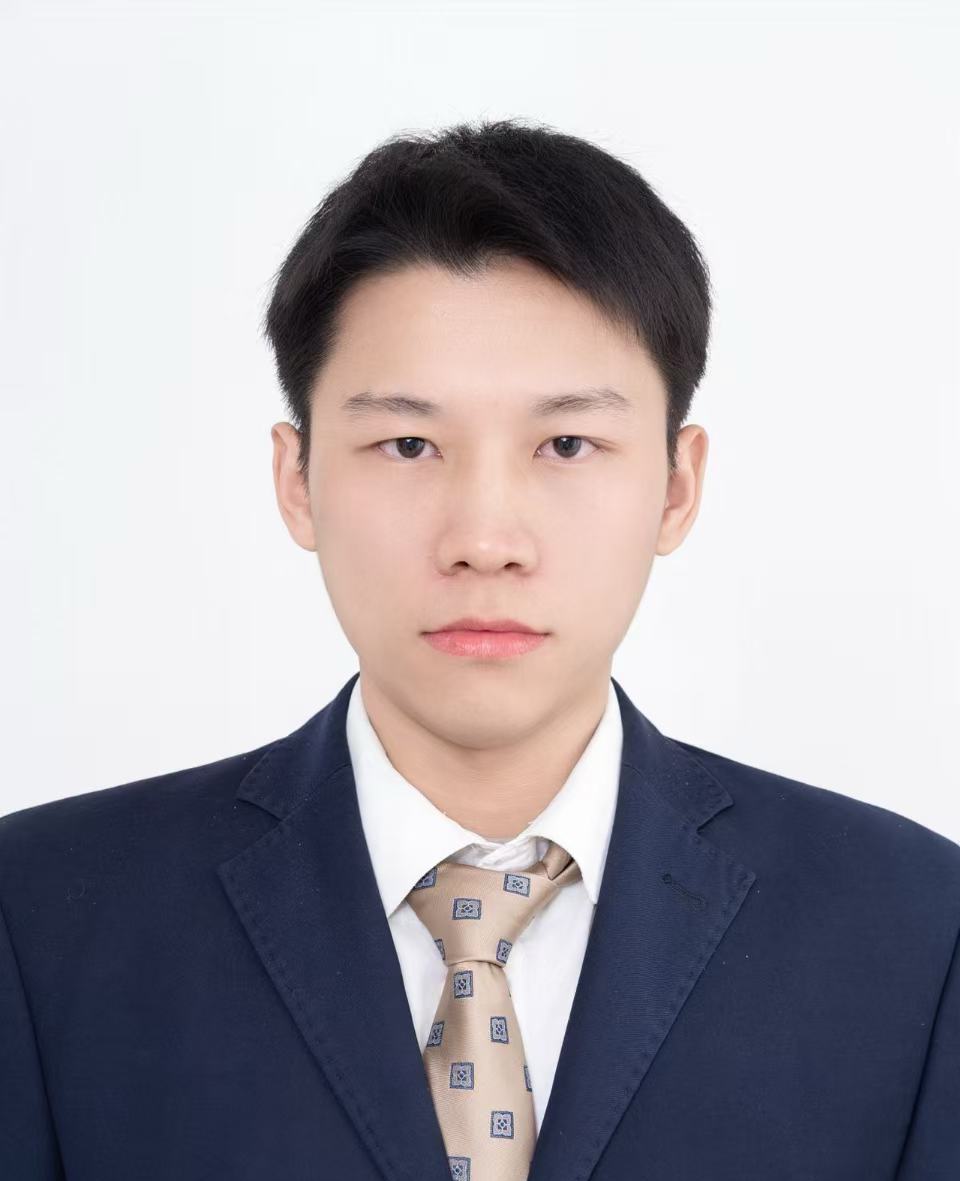}}]{Shuanghong Huang}
is currently a Ph.D. student at the School of Computer Science, Beijing Institute of Technology, China. His research interests include natural language processing, spoken language understanding, and large language models.
\end{IEEEbiography}
\bioskip

\begin{IEEEbiography}[{\biopic{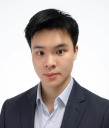}}]{Jiashen Liu}
is currently a Ph.D. student in Computer Science at the University of Warwick, United Kingdom. His research interests include artificial intelligence for engineering systems, machine learning, and interdisciplinary applications of AI in engineering.
\end{IEEEbiography}
\bioskip

\begin{IEEEbiography}[{\biopic{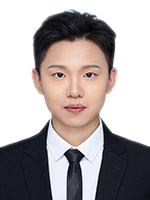}}]{Ming Lei}
is currently pursuing the master's degree at Harbin Engineering University, China, and is a research intern with Great Bay University, Dongguan, China. His research interests focus on artificial intelligence and its applications in engineering systems.
\end{IEEEbiography}
\bioskip

\begin{IEEEbiography}[{\biopic{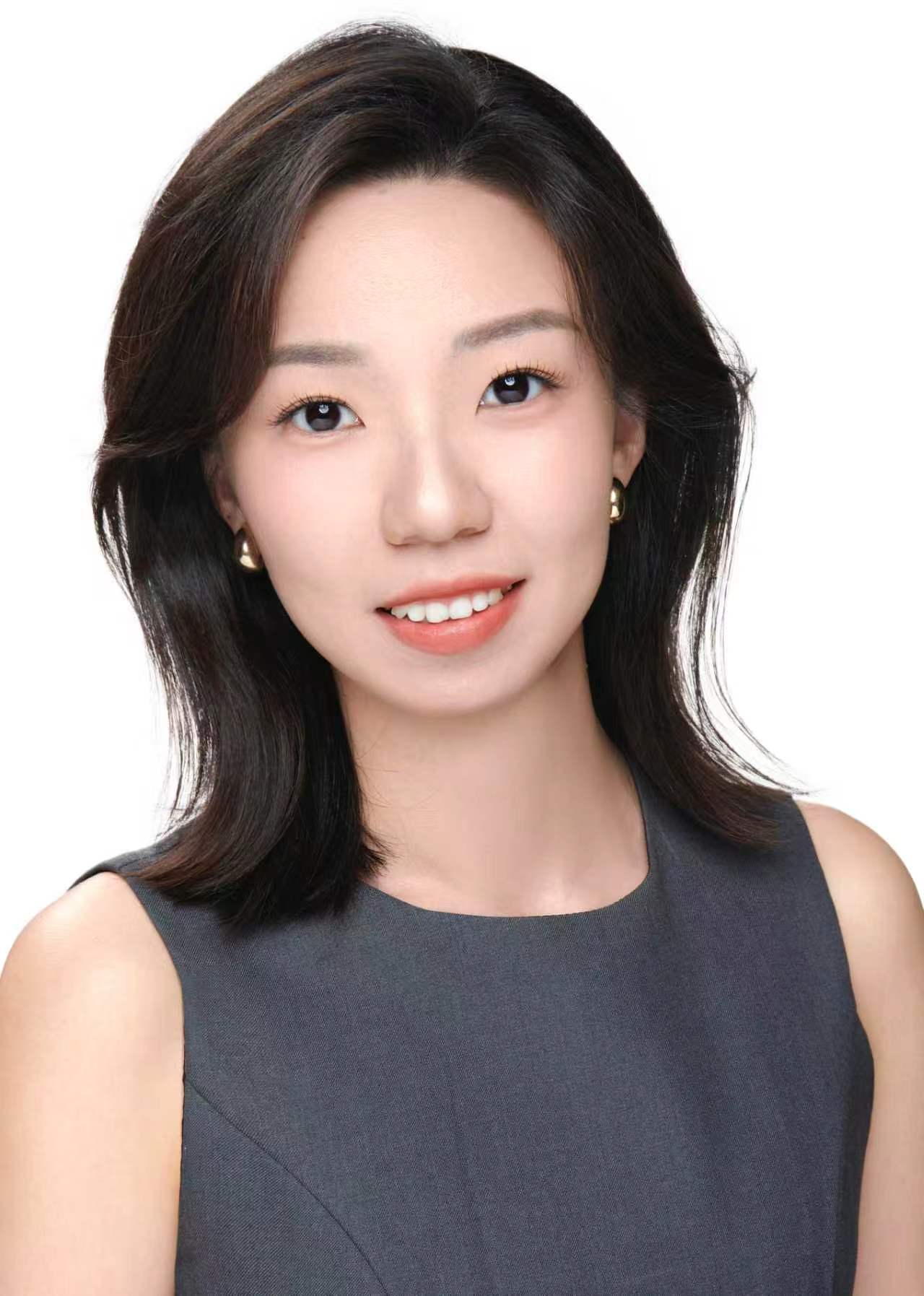}}]{Jidu Xing}
received the B.Eng. and Ph.D. degrees from the University of Hong Kong in 2017 and 2022, respectively. She is currently an Assistant Professor at the City University of Hong Kong (Dongguan), China. Her research interests include infrastructure failure analysis, smart hospital operation management, and applications of machine learning in systems engineering.
\end{IEEEbiography}
\bioskip

\begin{IEEEbiography}[{\biopic{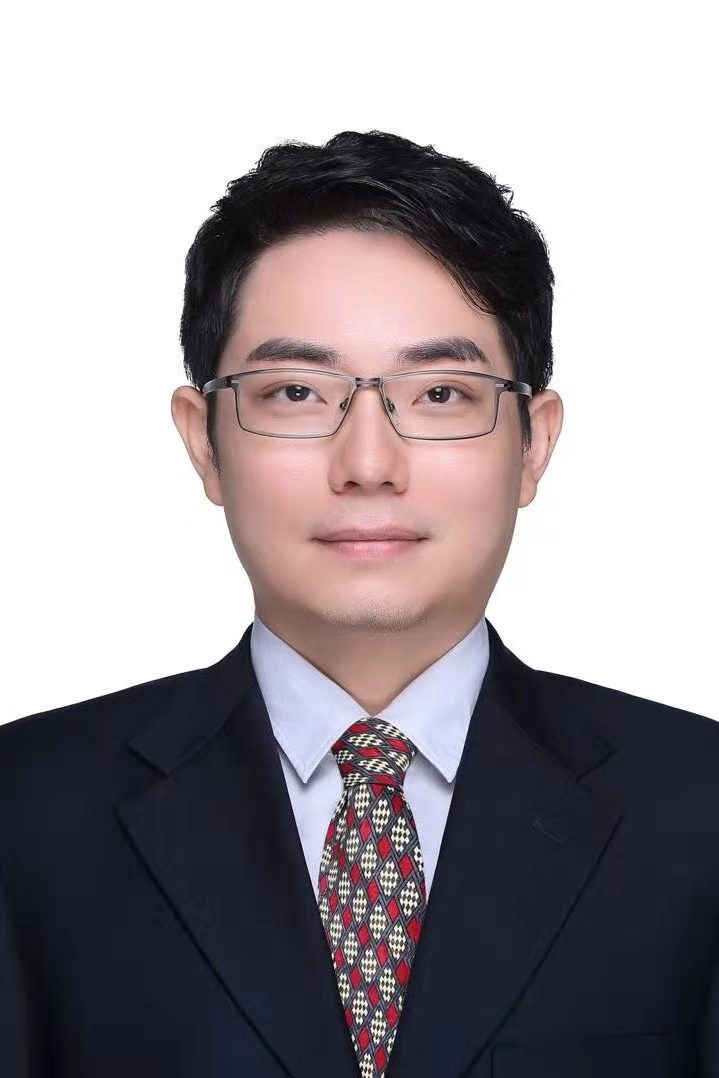}}]{Kaishun Wu}
(Fellow, IEEE) received the Ph.D. degree in computer science and engineering from the Hong Kong University of Science and Technology. He is currently the Associate Vice President for Research of the Hong Kong University of Science and Technology (Guangzhou), China, where he provides overall leadership for research development and management. He is also a Full Professor with the Data Science and Analytics Thrust and the Internet of Things Thrust under the Information Hub. Before joining HKUST(GZ), he was a Distinguished Professor and the Director of the Guangdong Provincial Wireless Big Data and Future Network Engineering Center at Shenzhen University. He has published more than 200 papers in major international journals and conferences and holds more than 100 invention patents, including nine U.S. patents. His research interests include wireless sensing, mobile computing, wireless big data, future networks, and intelligent Internet of Things. He has received numerous awards, including the 2012 Hong Kong Young Scientist Award, the 2014 Hong Kong ICT Awards: Best Innovation Award, the 2014 IEEE ComSoc Asia-Pacific Outstanding Young Researcher Award, and multiple provincial and national science and technology awards.
\end{IEEEbiography}
\bioskip

\begin{IEEEbiography}[{\biopic{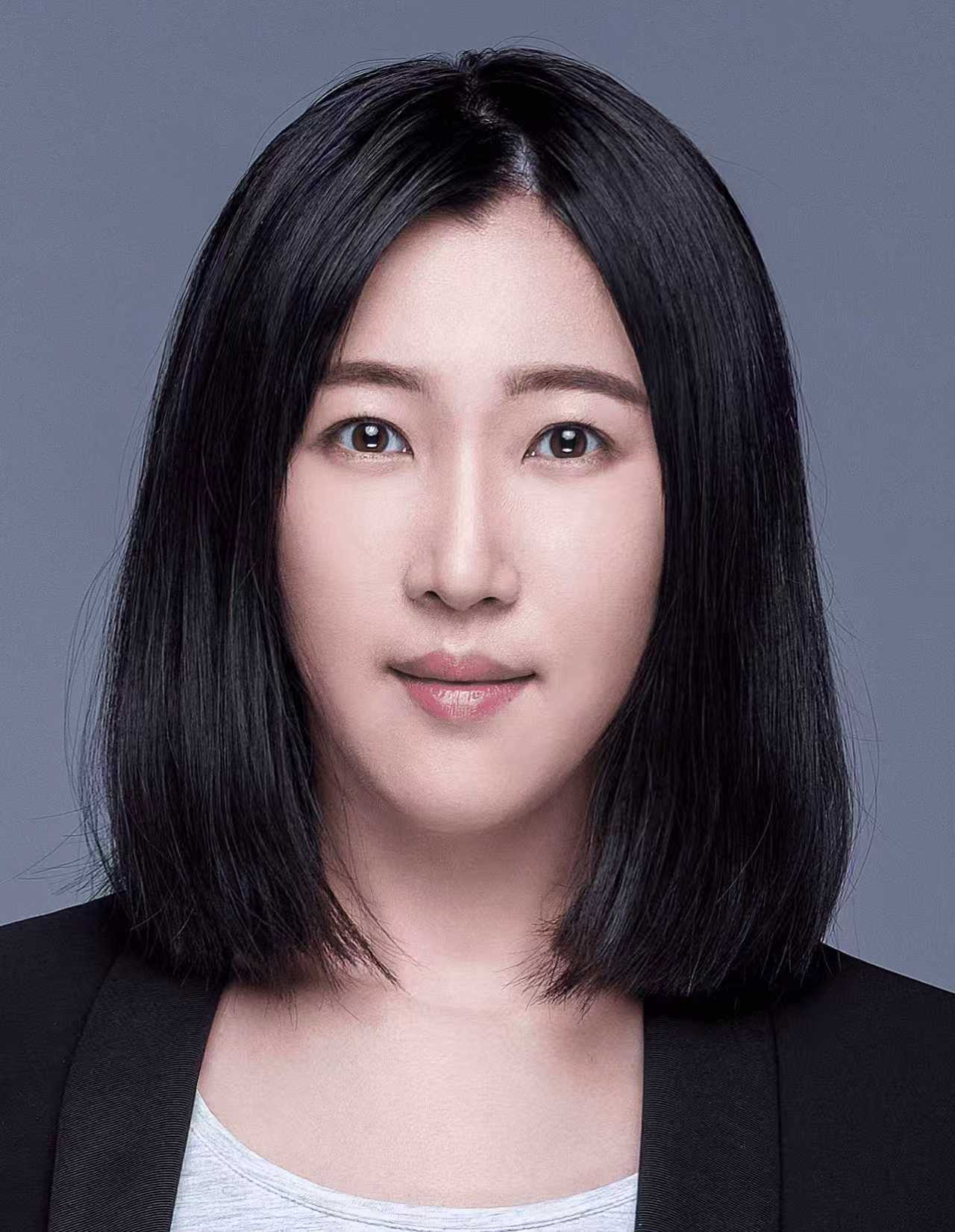}}]{Lu Wang}
(Senior Member, IEEE) is currently an Associate Professor with the College of Computer Science and Software Engineering, Shenzhen University, China. Her research interests include wireless communications, mobile computing, edge intelligence, and intelligent sensing systems.
\end{IEEEbiography}
\vspace{-3.5em}

\begin{IEEEbiography}[{\biopic{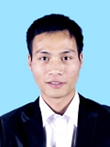}}]{Fei Luo}
is an Assistant Professor of Computer Science at Great Bay University, Dongguan, China. He received the Ph.D. degree in Electronic Engineering from Queen Mary University of London in 2020. His research interests include human activity recognition, wireless sensing, edge intelligence, and multimodal fusion. His work has been published in leading venues including IEEE Transactions on Mobile Computing.
\end{IEEEbiography}

\end{document}